\documentclass{article}
\usepackage{iclr2027_conference,times}
\usepackage[utf8]{inputenc}
\usepackage[T1]{fontenc}
\usepackage{courier}
\usepackage{amsmath,amssymb,amsfonts}
\usepackage{booktabs,tabularx,array,multirow}
\usepackage{graphicx,xcolor}
\usepackage{microtype}
\usepackage{enumitem}
\usepackage{listings}
\usepackage{placeins}
\usepackage{hyperref}
\usepackage{url}
\usepackage{xurl}
\usepackage{caption}
\hypersetup{colorlinks=true,linkcolor=black,citecolor=black,urlcolor=black,pdfauthor={Qinchuan Cheng, Zhantao Gong, Pengzhan Sun, Angela Yao, Shijie Li},pdftitle={Ego2World: Compiling Egocentric Video into Executable Planning Environments},pdfsubject={},pdfkeywords={}}
\iclrfinalcopy
\newcommand{\kb}{\textsc{Ego2World}}
\newcommand{\Gw}{G^{\mathrm{w}}}
\newcommand{\Gb}{G^{\mathrm{b}}}

\newcommand{\StrictTSR}{\ensuremath{\mathrm{Strict\text{-}TSR}}}
\newcommand{\GCR}{\ensuremath{\mathrm{GCR}}}
\newcommand{\RefTCR}{\ensuremath{\mathrm{Ref\text{-}TCR}}}

\newcommand{\indicator}{\mathbb{I}}
\newcommand{\True}{\textsc{true}}
\newcommand{\False}{\textsc{false}}
\newcommand{\Unknown}{\textsc{unknown}}

\title{\kb: Compiling Egocentric Video into Executable Planning Environments}
\author{%
Qinchuan Cheng$^{1}$ \quad Zhantao Gong$^{2}$ \quad Pengzhan Sun$^{3}$ \quad Angela Yao$^{3}$ \quad Shijie Li$^{4}$\\
$^{1}$Xi'an Jiaotong University (xjtu.edu.cn) \quad $^{2}$Nankai University (nankai.edu.cn)\\
$^{3}$National University of Singapore (nus.edu) \quad $^{4}$Institute for Infocomm Research, A*STAR (i2r.a-star.edu.sg)
}
\begin{document}
\maketitle
\fancyhead{}
\begin{abstract}
Egocentric videos capture how people carry out everyday activities, yet testing an agent requires evaluating the consequences of actions it chooses itself. We introduce \kb{}, a benchmark that turns annotated cooking activities into executable planning environments under partial observation. Its compiler links source steps and objects to symbolic action rules, persistent world states, and explicit task conditions, so researchers can execute an agent's proposed actions and check their outcomes. World state and agent belief are maintained separately, enabling controlled studies of planning and information reuse across continuing tasks. Across six planners on 105 tasks, Operational-TSR provides a strict end-to-end completion baseline, while condition-level attainment and execution diagnostics provide the primary discriminative comparison. Paired comparisons identify differences in action validity and execution completion after joint multiplicity adjustment. In a separate paired Qwen-Plus study, persistent belief improves action validity by 4.15 percentage points and reduces visual-query attempts by 90.27\%, with higher token use and no detected completion gain. \kb{} supports complementary comparisons of task attainment, execution behavior, and resource use, connecting recorded human activity to the development and evaluation of interactive agents.
\end{abstract}

\section{Introduction}
\label{sec:intro}

An agent preparing a drink must do more than name a plausible sequence of actions. It must find the ingredients, use the right container, and actually establish the required contents. Picking up the glass may be valid preparation even if nothing is ever poured into it. As work continues, earlier actions change what can happen next, and earlier observations may help a decision or become outdated. Evaluating such an agent requires tasks in which its choices have consequences, together with evidence of what those choices achieved.

Egocentric datasets such as EPIC-KITCHENS, Ego4D, and HD-EPIC capture the objects, interactions, and temporal structure of everyday work~\citep{damen2022rescaling,grauman2022ego4d,hdepic2025}. Video benchmarks use this evidence to assess procedural understanding and planning~\citep{jia2022egotaskqa,qiu2026egoplanbench2}, while interactive environments supply executable dynamics and goal-based evaluation~\citep{kolve2017ai2thor,shridhar2020alfred,li2023behavior}. Bringing a recorded activity into an interactive test requires connecting its objects and operations to rules that respond to new action proposals, while retaining the source evidence and defining what counts as completing the task. This motivates our question: \emph{how can an egocentric activity recording become a traceable, executable test of an agent's own decisions?}

\begin{figure}[t]
  \centering
  \includegraphics[width=\linewidth]{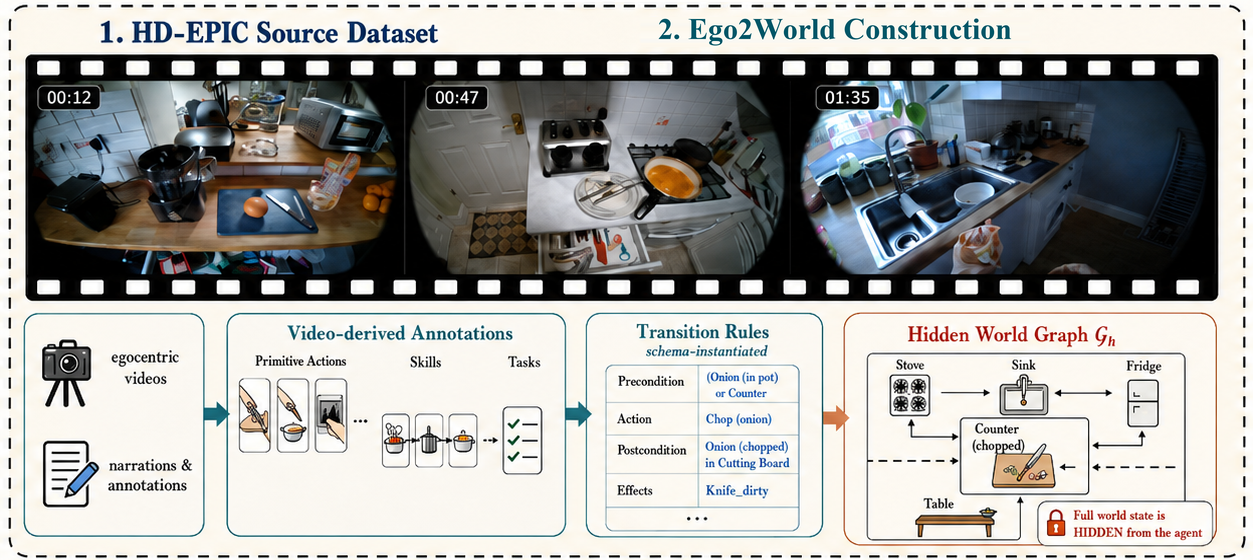}
  \par\smallskip
  {\small
  \begin{tabularx}{\linewidth}{@{}XcXcX@{}}
  \textbf{Agent decision} & $\longrightarrow$ & \textbf{World execution} & $\longrightarrow$ & \textbf{Task evaluation} \\
  Local information guides an action proposal. && Rules check the action and update the world. && Conditions check states or accepted events. \\
  \end{tabularx}
  \par\smallskip
  Local observations and execution feedback return to the agent for its next decision.}
  \caption{From activity evidence to interactive evaluation. The upper panel illustrates construction: its hidden graph $G_h$ is denoted $\Gw$ in the text, and skills correspond to action groups. The lower schematic shows how agents use the compiled environment. Source links, action records, and task checks make decisions inspectable; Table~\ref{tab:step_outcome_example} follows a real task through these stages.}
  \label{fig:overview}
\end{figure}

We introduce \kb{}, a benchmark that compiles HD-EPIC cooking annotations into episode worlds, executable operations, transition records, and task conditions (Figure~\ref{fig:overview}). Source-step links preserve the evidence behind objects and operations, while curated rules give new action proposals explicit consequences. A researcher can trace a task back to the activity that motivated it, run an agent in its compiled world, and examine whether its operations achieve the required outcome. The contribution is a reusable executable dataset that connects the construction of a task to the behavior of an agent attempting it.

The environment keeps the actual world separate from what the agent knows. The world contains object locations, relations, and states; the agent builds a partial record from local observations and action feedback. Historical video frames provide appearance and object context, and local symbolic changes communicate the effects of intervening actions. World state persists across related tasks, while agent information can be retained or reset independently. Researchers can therefore vary a planning or memory policy under common world rules and measure its consequences for execution, observation demand, and task attainment.

Our experiments demonstrate two uses of this resource. First, six planners on 105 tasks are evaluated with a strict Operational-TSR completion baseline and a complementary metric profile. The profile retains condition-level attainment and exposes accepted operations and execution completion. A paired analysis jointly accounting for 45 diagnostic contrasts identifies differences in action validity and execution completion, providing a discriminative comparison beyond the all-condition endpoint. Condition checks and execution records show what each metric captures. Second, a paired Qwen-Plus study on 51 terminal-state tasks tests the retention policy: persistent belief raises action validity by 4.15 percentage points and reduces visual-query attempts by 90.27\%, with greater token use and no detected completion gain. Continuous resource profiles and paired task examples connect these aggregate effects to the operations and observations behind them.

We contribute an executable benchmark with three connected elements:
\begin{itemize}[leftmargin=*,itemsep=2pt,topsep=3pt]
  \item \textbf{Activity data with executable meaning.} Source annotations are linked to typed worlds, 155 action types, transition records, and task conditions, preserving the connection between recorded evidence and operations an agent can perform.
  \item \textbf{Controlled interaction under partial information.} Continuing world state, local observations, and step-level feedback support experiments that vary agent decisions and information retention within the same environment rules.
  \item \textbf{Evaluation from operations to outcomes.} Condition checks, shared planner baselines, and paired memory results demonstrate how to inspect execution and measure the benefits and costs of information reuse.
\end{itemize}

\section{Related Work}
\label{sec:related}

\paragraph{From egocentric activity to planning evaluation.}
EPIC-KITCHENS~\citep{damen2018scaling,damen2022rescaling}, Ego4D~\citep{grauman2022ego4d}, and HD-EPIC~\citep{hdepic2025} document everyday activity from the participant's viewpoint; Ego4D Goal-Step~\citep{song2023ego4d} adds hierarchical procedural structure. EgoTaskQA~\citep{jia2022egotaskqa} evaluates action dependencies, effects, goals, and beliefs through video questions. EgoPlan-Bench2~\citep{qiu2026egoplanbench2} tests planning decisions from task progress and current visual observations across real-world scenarios. VidEgoThink~\citep{cheng2024videgothink} covers hierarchical planning, grounding, and reward-related reasoning, while ECBench~\citep{dang2025ecbench} broadens the evaluation of egocentric cognition. These benchmarks establish the value of activity evidence for reasoning about what an agent should do. \kb{} evaluates an agent's decisions by executing them in a persistent symbolic world and checking the resulting task outcomes.

\paragraph{Interactive environments and world modeling.}
ALFRED~\citep{shridhar2020alfred} evaluates instruction following and goal attainment in AI2-THOR; VirtualHome~\citep{puig2018virtualhome} executes household programs and studies their generation from language and video. BEHAVIOR-1K~\citep{li2023behavior} combines predicate-based activities with physical and visual simulation. ENACT~\citep{wang2025enact} uses BEHAVIOR-derived interaction sequences to test forward and inverse world modeling through observation and action reordering. \kb{} connects unscripted source activities to executable symbolic interpretations: an operation retains its source evidence, while rules determine the consequences of an agent's choices. This makes both task construction and new execution traces available for inspection.

\paragraph{Planning with feedback and retained experience.}
SayCan~\citep{ahn2022saycan} grounds skill selection in affordance estimates; Inner Monologue~\citep{huang2022inner} and SayPlan~\citep{rana2023sayplan} use environmental feedback to revise plans. Embodied Agent Interface~\citep{liu2024embodied} evaluates goal interpretation, decomposition, sequencing, and transition modeling. MEMORA~\citep{yu2026memora} forms persistent memory from egocentric experience and evaluates its use in retrospective reasoning and future planning, including a robot demonstration. \kb{} provides an environment for studying how retained information affects continuing execution. World state and agent belief can be controlled separately, and accepted operations are linked to frozen task conditions. Our paired study instantiates this capability with a simple retention intervention, measuring its execution benefits, observation savings, and context cost.

\section{Constructing the Executable Dataset}
\label{sec:benchmark}

The dataset connects recorded evidence to tasks an agent can attempt. HD-EPIC annotations and scene assets supply the objects and operations; curated rules specify how actions can change them. Compilation produces episode worlds, operations with source links, transition records, and fixed task conditions. These components let researchers inspect a task's interpretation and evaluate new decisions in its activity context. Table~\ref{tab:step_outcome_example} follows one such task from narration to outcome.

\paragraph{Organizing source evidence.}
HD-EPIC contains 41 hours of unscripted activity across nine kitchens and 69 recipes~\citep{hdepic2025}. We parse temporal annotations, object references, and verb--noun descriptions into ordered source steps. Temporally local steps are grouped around a central operation, with supporting setup or cleanup actions retained when grounded in nearby evidence. Schema-constrained language-model assistance handles grouping and reconciliation; each group retains its source-step identifiers and temporal span. These links preserve the provenance and context of an operation through subsequent normalization and execution.

\paragraph{Normalizing entities and operations.}
Source descriptions vary in granularity and wording. A manually constructed vocabulary maps more than 300 source verb classes to 155 executable action types. Objects are represented as instances with semantic categories, locations, containment relations, and symbolic attributes. Resolving an action therefore requires both an operation type and its bound entities: opening a cupboard and opening a bottle are not interchangeable simply because they share a verb. Grouped operations are aligned with activity boundaries to form tasks; related tasks retain their order within an episode.

\paragraph{Giving annotations executable meaning.}
Scene annotations initialize a typed world graph
$\Gw_0=(\mathcal V_0,\mathcal E_0,\mathcal A_0)$.
Nodes represent areas, objects, substances, and derived products; edges and attributes encode relations and states. Normalized actions are interpreted through curated preconditions and effects. Retrieval, for instance, resolves the requested object and checks location and access conditions before changing its position. The same rules apply to reference and agent-proposed operations, making object choice and execution order consequential. Video evidence supplies the scene and activity context; the rule base gives alternative actions a consistent executable interpretation.

\paragraph{Checking the compilation.}
Deterministic checks cover source-step assignments, temporal order, normalized fields, and structural consistency. Ambiguous cases receive localized semantic review. We manually inspected a 5\% sample of annotations across P01--P07, separately from model-assisted cross-review of task conditions. Compilation records pre- and post-operation graphs and execution results. For task validation, a reference policy proceeds continuously through each episode, using world access to resolve entities and supply repairs through the executor. Its state and action records provide evidence for checking task conditions (Appendix~\ref{app:construction}).

\paragraph{Tasks and evaluation cohorts.}
A primitive is one rule-checked action; an action group organizes a local operation and its contextual primitives. A task pairs an instruction with fixed conditions, called its \emph{contract}. Terminal-state contracts specify the required final world. Where the source evidence and represented attributes do not support a sufficient terminal contract, successful-event contracts specify accepted, entity-bound operations. These two forms of evidence retain distinct interpretations in evaluation. Conditions are frozen before planner evaluation, and world state carries over across tasks.

In \emph{Add Squash}, source steps 5--7 become a group for opening the bottle, pouring squash, and closing it. The normalized pour binds squash to the glass; the task's terminal condition checks the glass's contents. An accepted glass pickup is useful preparation, while an accepted pour can establish the required contents. The two recorded attempts in Table~\ref{tab:step_outcome_example} show how the benchmark connects this construction to execution and outcome evidence.

\begin{table}[t]
\centering\small
\caption{A real task from source evidence to agent outcome: \emph{Add Squash}, episode \nolinkurl{P02-20240209-231412}. The two attempts come from the paired memory study (Section~\ref{sec:memory_analysis}); both finish all four groups. The frozen condition checks contents, not the quantity mentioned in the narration. Appendix~\ref{app:source_example} gives the source mapping.}
\label{tab:step_outcome_example}
\setlength{\tabcolsep}{4pt}
\begin{tabularx}{\linewidth}{@{}lX@{}}
\toprule
Stage & Recorded evidence and executable interpretation \\
\midrule
Source activity & At 8.82--10.49\,s, narration 6 reads: ``Pour a small 1/20th measure of squash into the glass.'' \\
Compiled operation & Steps 5--7 form group 3. Its reference pour binds the substance squash and the destination glass. \\
Task condition & The final glass contents must include squash; missing contents evidence is \Unknown{}. \\
Persistent belief & Pour accepted; glass contents include squash. Condition true, task succeeds. \\
Per-group reset & Squash pickups rejected; glass pickup accepted; no pour attempted. Contents evidence missing, task fails. \\
\bottomrule
\end{tabularx}
\end{table}

We define the reported resource scope through two frozen evaluation suites. The planner suite contains 105 tasks, 18 complete episodes, four scenes, and 3,441 groups per planner. The memory suite contains 51 terminal-state tasks, 22 episodes, seven scenes, and 3,878 groups per condition. Their manifests specify the tasks and execution order; they are separately defined cohorts rather than random samples of HD-EPIC. The source scale above describes the underlying recordings. Appendix~\ref{app:artifacts} identifies the assets needed to run new agents and the contents of the accompanying reproduction package.

\section{Interaction and Step-Level Feedback}
\label{sec:simulator}

\subsection{Executing Actions in the Hidden World}
\label{sec:world_rules}
The environment owns the current world $\Gw_t$. A primitive binds an action type to concrete entities and, where required, a destination or tool. Rules check object state, containment, location, agent position, and hand or tool requirements. For a rule set $\mathcal R$, execution can be written as
\begin{equation}
 (\Gw_{t+1},f_t)=
 \begin{cases}
  (\mathcal T_{\mathcal R}(\Gw_t,a_t),\mathsf{accepted}),
       & \mathrm{pre}(a_t;\Gw_t)=\True,\\
  (\Gw_t,\mathsf{rejected}(c)), & \text{otherwise}.
 \end{cases}
 \label{eq:world_transition}
\end{equation}
Here $\mathcal T_{\mathcal R}$ applies the action effects and any associated symbolic process updates; $c$ reports the failed checks. Accepted operations can change locations, relations, attributes, or process state. A rejected primitive leaves the world unchanged. Earlier accepted steps in the same operation retain their effects, so failure does not undo a partially executed plan.

At each step, the executor checks a concrete proposal and returns either an accepted operation or the reason it failed. An accepted operation can still leave the task unfinished: opening a cupboard, for instance, need not retrieve the object inside. The evaluator checks the world and accepted events for the task outcome, while local feedback helps the agent choose its next action. These two views connect immediate execution quality to eventual task success.

\subsection{Observing a Changing Symbolic World}
\label{sec:observation}
An agent observes the world through local symbolic information, action feedback, and visual queries. A query for area $\ell$ pairs a recorded region image with a textual account of local changes:
\begin{equation}
 o_t^\ell=\left(I_{\mathrm{anchor}}^\ell,\;
 \operatorname{Render}\!\left[\operatorname{Diff}(G_0^\ell,G_t^\ell)\right]\right).
 \label{eq:delta_observation}
\end{equation}
Here $G_0^\ell$ is the region's state at episode initialization, and $G_t^\ell$ is its current state. The delta reports changes to selected location, storage, and appliance fields within that region. Video thus has two roles: its annotations ground the task's construction, and its recorded images supply visual evidence during interaction. Changes caused by the agent are conveyed by the current local delta.

When an image and a current delta conflict, the delta takes precedence during fusion. A removal from the queried area, for example, overrides the object's presence in the image without revealing its unobserved destination. Repeated queries reuse the image with updated local changes. Appendix~\ref{app:delta_interface} specifies the disclosed fields, and Appendix~\ref{app:vlm_protocol} describes how the evaluated agent selects images from episode assets and fallback sources.

\subsection{The Evaluated Agent and Its Belief}
\label{sec:planning_pipeline}
The environment separates its world state from the information retained by an agent. Our evaluated agent represents this information, $\Gb_t$, as a lightweight object/region store containing estimated locations, known container states, and evidence sources. Local observations initialize and refresh the store. For a group instruction $\tau_g$, the planner proposes actions from this belief, the current observation, and recent execution history:
\begin{equation}
 P_t=\pi_\theta(\tau_g,\Gb_t,o_t,h_t),\qquad
 \Gb_{t+1}=\mathcal U(\Gb_t,o_{t+1},f_t).
 \label{eq:belief_planning}
\end{equation}
The model parameters remain fixed. World state carries over across operations and tasks within each episode; belief retention is controlled separately. The store and its update rule $\mathcal U$ are agent choices, making the same task worlds available for comparing other information-use policies.

All compared planners use the same agent pipeline. The planner receives the current compiled group instruction, local reachable-action hints, and a short history of feedback and completed actions. Before retrieving an object, the controller may navigate to a remembered location or request visual evidence if its location is unknown. The executor checks proposed actions and may attempt a simulator-generated repair. If execution still fails, the planner revises its remaining sequence from the updated context, subject to the same replanning limit. Both repair steps and planner/controller steps count as attempted actions. This protocol evaluates feedback-based planning with common assistance and group-level instructions.

Successful actions update selected location and container beliefs, followed by integration of local observations. Retained information can guide later retrieval and avoid repeated discovery, but its usefulness depends on what is stored. In the logged coffee example, a fine-grained placement target is later replaced by a coarser region observation (Appendix~\ref{app:examples}). The memory experiment varies cross-group retention, including the context it makes available to later decisions; Appendices~\ref{app:protocol} and~\ref{app:implementation} specify the update rules.

Runtime records connect attempted actions, their source and acceptance status, group-end belief, and task outcomes. Appendix~\ref{app:artifacts} distinguishes these model traces from the compiler's full graph-transition records.

\section{From Execution Signals to Evaluation}
\label{sec:metrics}

The execution records answer three different questions: which reference action types were attempted, which concrete operations were accepted, and whether the task's conditions were met. Evaluating all three helps distinguish broad action overlap from valid execution and achieved outcomes. In \emph{Add Squash}, for example, retrieving the glass can be valid even when no squash is poured into it; the outcome requires the contents condition to hold.

\paragraph{Outcome evidence.}
Each task $\tau$ has one frozen contract with a nonempty condition list $\mathcal C_\tau$. Terminal-state conditions read the hidden world at the task endpoint $T$; successful-event conditions read the accepted task-local execution history $H_\tau$:
\begin{equation}
 C_\tau(\Gw_T,H_\tau)=
 \begin{cases}
   \Phi_\tau(\Gw_T), & \text{terminal-state contract},\\
   \Psi_\tau(H_\tau), & \text{successful-event contract}.
 \end{cases}
 \label{eq:goal_contract}
\end{equation}
Conditions include their required entity bindings. Their truth values are $\{\True,\False,\Unknown\}$, with missing entities, attributes, or evidence treated as \Unknown{} and not satisfied. Planner declarations do not count as evidence. Terminal conditions allow different executable trajectories to the same required outcome. Event conditions accept specified action alternatives and bindings with minimum counts; they check performed operations without requiring the complete reference sequence.

\paragraph{Task success and partial attainment.}
Let $E_\tau(c)$ denote a condition's evaluation and $d_\tau$ indicate that the run has completed. Define
\begin{equation}
 s_\tau=d_\tau\,\indicator\!\left[\bigwedge_{c\in\mathcal C_\tau}E_\tau(c)=\True\right],
 \qquad
 \StrictTSR=\frac{1}{|\mathcal T_{\rm term}|}
             \sum_{\tau\in\mathcal T_{\rm term}}s_\tau.
 \label{eq:strict_success}
\end{equation}
Strict-TSR uses only terminal-state tasks. The 105-task comparison contains 34 such tasks and 71 successful-event tasks, and reports
\begin{equation}
 \OpTSR@105=\frac{1}{105}\sum_{\tau\in\mathcal T}s_\tau.
 \label{eq:operational_success}
\end{equation}
This operational rate preserves the fixed denominator and serves as the strict end-to-end completion baseline. We report contract strata for terminal and accepted-event attainment. Parse failures, empty plans, and execution blocks remain failures. \emph{Completed} means that the queue has finished; success additionally requires every condition. The conjunctive baseline is complemented by condition-level and execution diagnostics for model discrimination.

Goal completion ratio measures the fraction of frozen conditions satisfied:
\begin{equation}
 \GCR(\tau)=\frac{d_\tau}{|\mathcal C_\tau|}
    \sum_{c\in\mathcal C_\tau}\indicator[E_\tau(c)=\True].
 \label{eq:gcr}
\end{equation}
For a completed run, satisfying three of four conditions gives GCR $0.75$ and binary success zero. Task-macro averaging gives each task equal weight; it measures the fraction of its stated requirements met, without assuming equal predicate difficulty. Non-completed runs receive zero under the failure policy, not an observed all-false condition state. The 51-task memory study uses terminal conditions throughout and has its own denominator.

All planner diagnostics are task-macro means over the same 105 task IDs. GCR normalizes conditions within each task, while validity and completion contribute one value per task, so model differences are paired on an identical task distribution. Terminal and event contracts remain distinct reported strata.

\paragraph{Execution validity and reference coverage.}
Action validity is the fraction of attempts accepted by the executor within each task. It includes repairs and planner/controller actions; a validation check without execution is not an attempt. We report the unweighted mean of task-level fractions, using $\max(1,N_{\mathrm{attempted}})$ as each denominator.

Reference task coverage (Ref-TCR) measures attempted reference action types; it ignores order, binding, and acceptance (Appendix~\ref{app:diagnostic_aggregation}). Model metrics use macro averaging, while trace summaries count attempts. The capability profile uses strict completion, GCR, action validity, and completed-run rate for distinct attainment and execution axes. These dimensions remain separate; unresolved paired intervals remain ties on that axis, yielding a useful partial ranking without collapsing the metrics.

\section{Experiments}
\label{sec:experiments}

We use the benchmark to study how agent decisions translate into achieved outcomes and how retaining information changes execution. The six-planner comparison connects goal attainment to execution records; the paired memory study measures information reuse within continuing episode worlds.

\subsection{Shared Evaluation Protocol}
\label{sec:exp_setup}

The planner comparison uses the first 18 complete episodes in Scenes~1--4, with 105 frozen tasks and 3,441 groups per planner. Admission requires reference evidence, contract validation, semantic review, and a non-triviality check before planner execution. Ninety-eight tasks have explicit evidence coverage; seven event contracts are proxy-limited where reference evidence does not cover every instruction component. All remain in the primary denominator.

We evaluate Qwen-Plus, GPT-5.5, Gemini~3.1 Pro, DeepSeek-V4-Flash, GLM-5, and Claude Sonnet~4.6 in the shared pipeline of Section~\ref{sec:planning_pipeline}, including group instructions, controller navigation, and executor repair. Task order, episode initial worlds, Qwen-VL-Plus observation interface, persistent-memory policy, prompt structure, action interface, execution limits, and scoring rules are held fixed. Worlds then evolve with each agent's actions. Each planner uses its API's non-thinking request setting; provider adapters normalize response syntax before the common executor (Appendix~\ref{app:implementation}). Non-completed runs remain failures in the fixed denominator.

Pointwise confidence intervals use 10,000 percentile bootstrap samples (seed 42), resampling episodes within scenes to retain within-episode dependence. Paired differences reuse the sampled clusters; diagnostic comparisons additionally use simultaneous intervals (Appendix~\ref{app:statistics}). Intervals describe episode variation conditional on recorded trajectories, not model-run stochasticity. The memory study uses its separate matched cohort and the same cluster procedure.

\subsection{Task Evidence and Outcome Checks}
\label{sec:construction_validation}

Reference preparation executes source-derived operations in episode order, with one world reset at the episode start. It uses privileged entity binding and executor-checked repair actions, retaining task states and action records. The frozen protocol checks these records against terminal conditions or selected successful events. All 105 task IDs match: reference evidence satisfies every contract, while each recorded task-initial state fails its contract. All 317 predicate-level negative suites pass the expected rejection checks. A separate 12-task stress test covers 43 predicates and 45 replayed evidence changes; reference replay passes, and partial or wrong-binding cases fail. Both legal orders pass in all ten tasks permitting an independent update swap. These checks support the executable interpretation and selectivity of the task conditions. Appendix~\ref{app:evaluator_validation} details the continuous reference process, subsequent event-boundary adjustments, and the distinction between this construction evidence and an independently scored online oracle.

\subsection{Goal Attainment and Execution Profiles}
\label{sec:full_planner_comparison}

\begin{figure}[t]
\centering
\includegraphics[width=\linewidth]{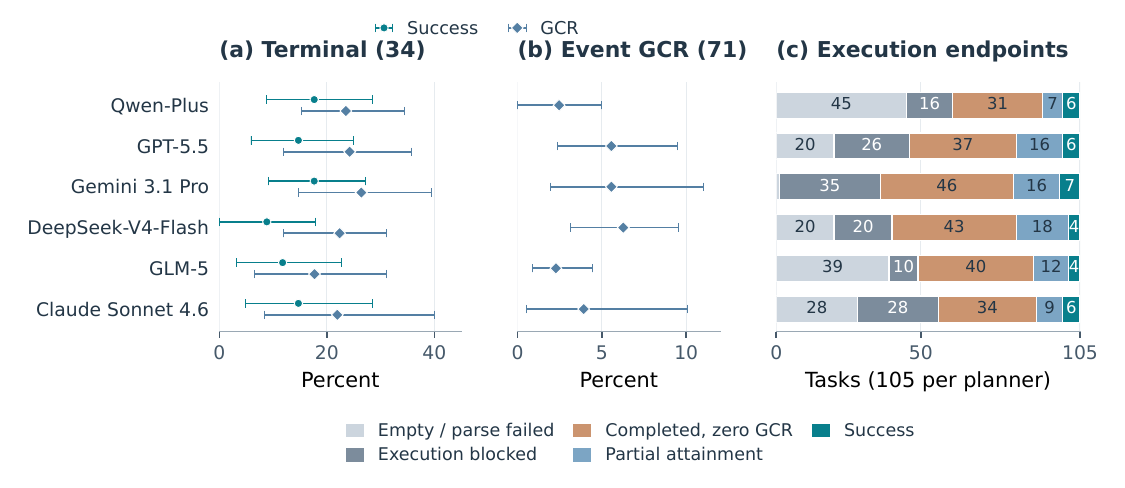}
\caption{Goal attainment within each contract type and execution endpoints for all 105 tasks. (a) Terminal Strict-TSR and GCR; (b) graded attainment (GCR) for successful-event contracts; both use pointwise 95\% episode-cluster intervals. Operational-TSR remains the fixed-denominator completion baseline in Table~\ref{tab:planner_results}. Percentage axes have different ranges. (c) Mutually exclusive recorded endpoints: partial attainment is $0<\GCR<1$ in a completed run. Counts describe outcomes, not their causes. Tables~\ref{tab:planner_results} and~\ref{tab:contract_strata} give aggregate and stratified values.}
\label{fig:planner_overview}
\end{figure}

Operational-TSR requires a finished execution and every required condition, and serves as the strict completion baseline. Figure~\ref{fig:planner_overview} places this endpoint beside graded progress and execution behavior: terminal-task Strict-TSR and GCR are 8.82\%--17.65\% and 17.70\%--26.40\%, while event-task GCR is 2.28\%--6.27\%. Because every planner is evaluated on the same task IDs with the same task-macro aggregation, these values support paired model comparisons; the terminal and event panels retain their distinct contract interpretations. The primary comparison is therefore a capability profile, not a pooled scalar score (Appendix~\ref{app:statistics}).

The all-condition baseline gives no partial credit by design: a completed partial trajectory shares its binary endpoint with a zero-attainment trajectory. Across the 630 records, 78 completed runs attain some but not all conditions. GCR preserves this graded distinction under the same contracts (Appendix~\ref{app:condition_structure}); excluding seven proxy-limited tasks preserves the complete-attainment counts at 4.08\%--7.14\% on 98 tasks.

The profile separates the intended axes (Figure~\ref{fig:planner_overview}c): Gemini leads GCR and completion, GPT-5.5 leads validity, and DeepSeek is high on event GCR. These are capability-specific orderings. Across 45 paired contrasts, simultaneous 95\% intervals identify GPT-5.5 over GLM-5 in validity (+8.61 pp, $[2.14,15.09]$) and four completed-run differences, including Gemini over Qwen-Plus (+23.81 pp, $[8.55,39.07]$). GCR preserves partial trajectories that a conjunctive endpoint collapses, making the profile useful alongside the strict baseline. The 322,748 attempts remain inspectable (Appendix~\ref{app:trace_analysis}).

\FloatBarrier
\subsection{Persistent Belief in Continuous Execution}
\label{sec:memory_analysis}

World--belief separation lets us vary information retention while preserving the execution setting. Our paired Qwen-Plus study uses 51 terminal tasks from 22 episodes and 3,878 groups per condition. Both conditions execute episodes continuously: persistent belief resets once per episode, whereas the comparison resets belief at each group. Task IDs, episode order, initial worlds, planner, VLM, prompts, budgets, code versions, and evaluator are paired.

\begin{figure}[t]
\centering
\includegraphics[width=\linewidth]{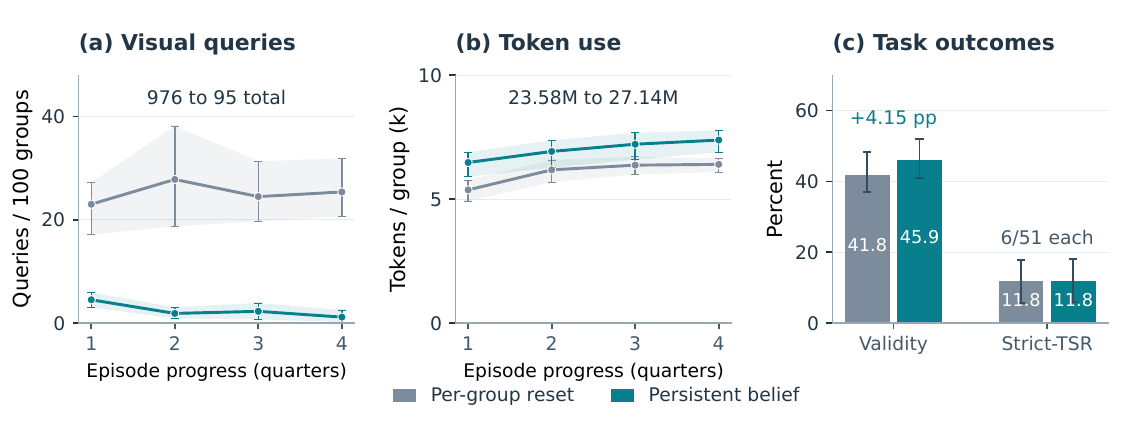}
\caption{Paired memory study on 51 tasks. (a,b) Post-hoc resource profiles across four quarters of each episode's group sequence; all 22 matched episodes contribute to every quarter. Rates pool groups, while (c) averages task metrics. Whiskers and shading show pointwise 95\% episode-cluster intervals. The validity gain has paired CI $[+0.11,+7.78]$ pp; no completion gain is detected. Full paired effects and aggregation details appear in Appendix~\ref{app:memory_continuity}.}
\label{fig:memory_continuity}
\end{figure}

Retaining belief raises task-mean action validity from 41.79\% to 45.94\%, a $+4.15$ percentage-point (pp) difference with 95\% CI $[+0.11,+7.78]$~pp. Visual-query attempts fall from 976 to 95, a reduction of 90.27\% (CI $[87.62\%,92.97\%]$). This reduction comes with higher token use: group-accounted tokens rise from 23.58M to 27.14M, a $+3.56$M difference with CI $[+1.82\mathrm{M},+5.45\mathrm{M}]$. Both conditions attain 6/51 terminal successes; the paired difference is 0.00~pp with CI $[-6.38,+6.25]$~pp. The retention policy therefore improves execution validity and reduces fresh observation demand, while increasing the context processed by the agent. These are complementary effects of retaining information, with no detected completion gain.

Figure~\ref{fig:memory_continuity} follows resource use through the continuing episodes. We divide each episode's ordered groups into four progress quarters, retaining the same episodes and source groups in both conditions. Persistent belief has lower query demand and higher token use in every quarter. The profiles show that the aggregate trade-off extends across the execution sequence; the controller can reuse retained locations when deciding whether to query again. These are descriptive profiles through the same episodes; task difficulty and planning horizon are not separately varied.

The task records explain why improved local validity does not guarantee a completed goal. In \emph{Add Squash}, both conditions finish all four groups with similar validity, but only persistence produces the accepted pour that establishes the contents condition (Table~\ref{tab:step_outcome_example}). In \emph{Prepare mushrooms}, persistence accepts 85/122 attempts yet reaches an execution block; reset accepts 26/46 and finishes with the required clean state. Four tasks succeed in both conditions, two only with persistence, and two only with reset. Appendix~\ref{app:paired_cases} includes all four discordant cases. Together, the paired intervention and operation records demonstrate how the benchmark measures policy effects and locates the execution differences accompanying task outcomes.

\FloatBarrier
\section{Discussion and Limitations}
\label{sec:discussion}

\paragraph{A benchmark for agent development.}
Source links explain construction; executable rules make choices consequential; outcome checks show achieved states. The strict baseline anchors end-to-end attainment, while task-normalized graded progress and execution diagnostics provide comparable, interpretable capability axes for development. The planner and memory studies support an inspect, revise, and recompare cycle.

\paragraph{Scope and open directions.}
The curated rules cover symbolic cooking activities; extending the domain requires new ontologies, evidence, and action semantics. Checks establish source correspondence and consistency within scope. The interface combines historical anchors, symbolic changes, instructions, and execution assistance; broader cohorts and matched isolation are natural extensions. Model comparisons and memory retention do not separately isolate planning horizon. Memory findings concern Qwen-Plus.

\section{Conclusion}
\label{sec:conclusion}

\kb{} links egocentric evidence, executable worlds, agent decisions, and outcome checks; its strict, graded, and execution metrics provide actionable capability axes for planning and memory studies.

\FloatBarrier
\label{page:main_end}
\clearpage
\section*{Ethics Statement}
\kb{} builds an executable annotation layer over HD-EPIC and adds no new human-subject recordings. Source videos remain subject to the dataset's access terms; the manuscript source package contains neither raw videos nor API credentials. Kitchen, activity, and cultural coverage are limited by the source collection. Symbolic action acceptance is not a certification of physical safety, and these results do not establish deployment readiness for kitchen robots.

\section*{Reproducibility Statement}
Appendices~\ref{app:construction}--\ref{app:metrics} specify construction, interaction, and scoring; Appendix~\ref{app:implementation} records model settings and runtime behavior. The accompanying manuscript source package includes anonymized frozen exports, task-level metrics, hashes, and table-regeneration scripts, with scope documented in Appendix~\ref{app:artifacts}. These artifacts reproduce the reported tables and figures from frozen evidence, including the contract-type analysis and all 45 post-hoc diagnostic contrasts. The separate benchmark release will provide the compiler, executor, rules, initial worlds, source mappings, and frozen task contracts for the two evaluation suites. Full agent re-execution additionally requires the applicable source-data and provider access; Appendix~\ref{app:artifacts} distinguishes these components from manuscript reproduction.

\section*{AI Use Statement}
Generative AI assisted annotation grouping, reconciliation, and goal review, and supplied the evaluated planner and visual-query components. Manuscript assistance covered scientific critique, interpretation and offline analysis of records, code and evidence checks, drafting, editing, references, and validation scripts. Results and examples were checked against archived experimental records. The method figures preserve original artwork with text corrections and cropping; statistical figures are plotted from archived records. No generative image editing was used. The authors remain responsible for all claims, evidence, citations, and disclosures.

\clearpage
\bibliographystyle{iclr2027_conference}
\bibliography{references}
\clearpage
\appendix
\section{Annotation-to-World Construction}
\label{app:construction}

\begin{figure}[htbp]
  \centering
  \includegraphics[width=\linewidth]{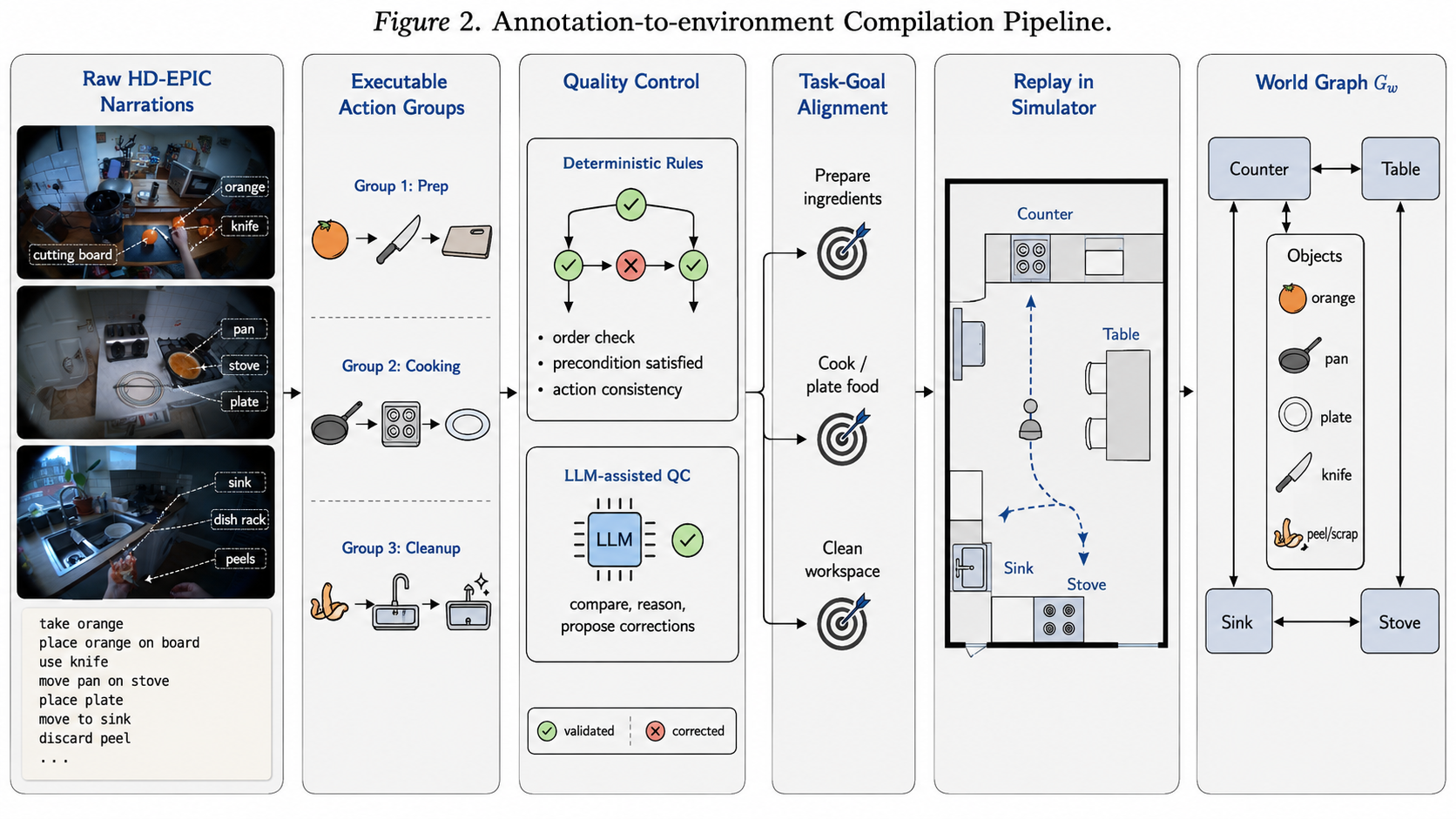}
  \caption{Annotation-to-environment compilation pipeline. Raw HD-EPIC narrations are grouped into executable action groups, validated through deterministic and LLM-assisted quality control, aligned with task goals, and replayed in the simulator to produce graph-transition worlds.}
  \label{fig:annotation_pipeline}
\end{figure}

The executable dataset joins source evidence, normalized operations, world
transitions, and task conditions. Figure~\ref{fig:annotation_pipeline} shows
how the construction preserves these connections from narrations to
executable graph-transition samples.

\subsection{Source Evidence and Normalization}
\kb{} adds an executable layer to HD-EPIC~\citep{hdepic2025}; the underlying
recordings remain real observations of unscripted cooking. The source
annotations provide temporal bounds, verb--noun descriptions, manipulated
objects, and activity structure. They constrain which entities and
interactions enter the compiled world, while a normalized ontology makes
these observations usable by the symbolic executor. The compilation therefore
has two distinct responsibilities: preserve the evidence of what occurred,
and express supported operations in a consistent executable vocabulary.

For each video, we parse the official narrations into a manifest with temporal
bounds, verb--noun labels, keyframe indices, and storage-object indicators.
Overlapping temporal windows make the annotation sequence manageable for
schema-constrained LLM grouping. Each output group records its covered source
steps, temporal span, manipulated objects, normalized primitives, contextual
actions, and state-change descriptions. Reconciliation joins window outputs
into a single ordered sequence, recovers uncovered source-step assignments,
and recomputes group boundaries. A contextual action such as closing a storage
or turning off an appliance is retained when supported by nearby evidence.
Source-step assignments make these contextual choices inspectable.

The manually constructed normalization maps more than 300 source verb classes
to 155 executable action types. This separates lexical variation from the
operations the simulator can validate. The same action type can occur with
different object instances and preconditions, so vocabulary size is not the
number of concrete transitions. Instance identities, containment, symbolic
state, and functional-area membership remain relevant after normalization.

\subsection{Structural and Semantic Quality Control}
A deterministic validator checks source-step coverage, normalized hand labels,
storage interactions, state-change descriptors, and abnormal group structure.
High-confidence structural repairs are applied automatically. Uncertain
cases receive localized semantic review rather than unrestricted regeneration.
LLM-assisted revisions must retain their source-step support, preserve temporal
order, and satisfy the executable schema. We manually inspected a 5\% sample
of annotations across P01--P07 as part of quality control. The manual sample inspection complements automated structural checks and
localized semantic review, keeping accepted annotations tied to specific
evidence and executable constraints.

Validated groups are merged at the participant-scene level and aligned to
high-level activities using temporal overlap and proximity. Goal-aligned,
temporally contiguous groups become task units. The world graph is initialized
from the annotated scene; the transition rules then define how candidate
actions change it. Human normalization and curated symbolic semantics provide the executable
interpretation, while source videos and annotations provide the activity
evidence. This explicit separation makes both the provenance of an operation
and the rule governing its consequences available for inspection.

\subsection{Replay and Frozen Task Admission}
For group $g_t$, compilation expands its normalized operation into primitives
$a_t^{1:K}$ and records execution against the hidden world:
\begin{equation}
 (\Gw_t,g_t)\longmapsto
 (a_t^{1:K},\Gw_{t+1},\Delta_t,\xi_t),
 \label{eq:simulator_compilation}
\end{equation}
where $\Delta_t$ records graph changes and $\xi_t$ records execution metadata,
including success, invalidity, or skipped actions. These records expose whether
an annotated operation has a coherent executable interpretation. The
transition collection is
\begin{equation}
 \mathcal D_{\mathrm{trans}}=
 \{(\Gw_t,a_t^{1:K},\Gw_{t+1},\Delta_t,\xi_t)\}_t.
 \label{eq:transition_dataset}
\end{equation}
Each transition sample connects an operation to its input state, effects,
and execution status. These records support transition inspection and
comparison with agent behavior. Full pre- and post-states belong to the
compiler and evaluator; agent observations follow the separate disclosure
protocol in Appendix~\ref{app:protocol}.

Task admission to the controlled evaluation requires reference replay, goal
validation, semantic review, and a non-trivial predicate check. These checks
are completed before locking the task contracts for planner evaluation.
Goal preparation uses model-assisted primary annotation, independent
cross-review, and adjudication; these records are distinct from the manual
annotation sample check described above.
Necessary terminal predicates specify the outcome and its object bindings;
irrelevant agreement with all attributes of the reference endpoint is not
required. Tasks scored through event contracts retain their separately
specified operational criteria. One goal variant is enabled for each task in
the current evaluation. Each contract is frozen independently of later planner
outcomes. Reference preparation resets the world once per episode and executes
source operations in group order. It uses world access to resolve entity names
and to propose preparatory actions, such as navigating to a container,
opening it, or releasing a hand. These actions pass through the executor;
the process retains action checks, repair outcomes, state hashes, and task
start and end graphs. This continuous execution supplies the reference
evidence used during goal preparation. Appendix~\ref{app:reference_evidence}
explains how later event-contract preparation selects evidence and adjusts
task boundaries, and how the checks relate to planner evaluation.

Replay validation, semantic review, and alternative-order tests answer
different questions. Replay records the states and accepted events produced
by reference execution; semantic review examines their support in the
activity evidence. Alternative-order tests check that independent operations
can be exchanged without changing the required outcome. The 5\% manual
annotation inspection complements these checks, but no quantitative human
error rate or inter-annotator agreement is estimated from that inspection.

\begin{table}[htbp]
\centering
\small
\caption{Source-data context and the two evaluation cohorts. HD-EPIC scale
is not the evaluated scope. The six-planner cohort is a non-random prefix of
complete episodes; the memory cohort is defined separately.}
\label{tab:construction_scope}
\begin{tabular}{lr}
\toprule
Property & Count \\
\midrule
Source recordings & 41 hours; 9 kitchens; 69 recipes \\
Normalized executable action types & 155 \\
\midrule
Controlled task set & 105 \\
Complete episodes in controlled set & 18 \\
Scenes in controlled set & 4 \\
Execution groups per planner & 3,441 \\
Tasks with terminal-state contracts & 34 \\
Tasks with success-event contracts & 71 \\
\midrule
Paired memory tasks (all terminal-state) & 51 \\
Complete episodes in memory cohort & 22 \\
Scenes in memory cohort & 7 \\
Execution groups per memory condition & 3,878 \\
\bottomrule
\end{tabular}
\end{table}

\FloatBarrier
\subsection{A Source-to-Outcome Example}
\label{app:source_example}
Table~\ref{tab:step_outcome_example} follows \emph{Add Squash} in episode
\nolinkurl{P02-20240209-231412}. Its source narration manifest records step~6
at 8.82--10.49\,s: ``Pour a small 1/20th measure of squash into the glass.''
The preceding step describes unscrewing the cap and the following step
describes closing it. The checked compilation groups steps 5--7 into
group~3, spanning 7.52--12.16\,s, with the instruction ``Open the bottle,
pour squash into the glass, and close the bottle.'' The source noun label
for step~6 is \texttt{spoon}; the narration identifies squash and the glass,
which the checked group retains as the substance and destination.

The four compiled group instructions match the frozen inputs used in the
paired evaluation. The normalized reference operation is
\texttt{pour(object=squash, target=glass)}. Its associated source evidence
describes squash moving from the bottle into the glass; the frozen terminal
condition tests whether \texttt{glass.state\_flags.contains} includes
\texttt{squash}. This condition checks the transfer outcome. It does not
score the narrated measure, bottle-cap state, or glass placement.

The paired execution records then show how this task is attempted. Both
conditions retrieve squash in group~2. With persistence, group~3 contains
an accepted pour and the terminal contents condition is true. With per-group
reset, three squash pickups are rejected, a glass pickup is accepted, and
no pour is attempted; the required contents attribute is missing and the
condition is \Unknown{}. All four groups finish in both runs. Thus source
evidence, operation binding, action acceptance, and outcome judgment can be
examined together without inferring success from a completed queue.
Appendix~\ref{app:paired_cases} provides the attempt counts and paired
outcomes. The package's \texttt{data/benchmark\_running\_example.json}
retains the selected source records, normalized actions, goal condition,
and file hashes. This is an offline correspondence check of existing
records, not an additional agent run or an alternative-order experiment.

\subsection{What Transfers to Another Domain}
The separation of world state from belief, restricted observations, explicit
queries, precondition--effect execution, and contract-based scoring does not
depend on cooking. A new domain still requires an appropriate ontology,
instance annotations, action vocabulary, transition-rule instances, visual
anchors, and goal definitions. It also requires sufficiently detailed temporal
evidence for normalization and replay. The current dataset instantiates these components in kitchen activities.
Its construction interfaces identify the data and semantics needed for
extension; cross-domain evaluation remains future work.
Table~\ref{tab:construction_scope} reports the source context and the two
controlled evaluation cohorts.

\section{World Rules, Observations, and the Agent Protocol}
\label{app:protocol}

\subsection{Primitive-Level Rule Semantics}
\label{app:rules}
A primitive action binds arguments to typed entities. Its rule tests atomic
predicates such as $\operatorname{at}(o,\ell)$,
$\operatorname{state}(o,s)$, and $\operatorname{contains}(c,o)$:
\begin{equation}
 \mathrm{pre}(r)=\bigwedge_j \phi_j(\mathcal V_t,\mathcal E_t,\mathcal A_t).
\end{equation}
The effect comprises graph mutations
\begin{equation}
 \mathrm{eff}(r)=
 (\Delta_V^+,\Delta_V^-,\Delta_E^+,\Delta_E^-,\Delta_A),
\end{equation}
where the first four terms add or remove nodes and edges and $\Delta_A$
updates attributes. Effects are applied only when the preconditions hold.
Otherwise the failed primitive leaves its input graph unchanged and returns
the violated precondition. This is a primitive-level guarantee: any earlier
successful primitive in a multi-action operation has its own valid state
transition. No task success can be obtained by merely proposing an invalid
state change.

\subsection{Information Available to the Agent}
Separating world state from agent information makes observation and retention
policies explicit experimental controls. The executor owns $\Gw$ and the rule checks. The agent receives a compiled group
instruction, the permitted action schema, local observations, and execution
feedback. Its planner context is assembled from agent-side
information: current position and hand state, known objects, completed-action
summaries, received state deltas, and failure messages. Local observations also include reachable-action hints;
the executor can supply repair primitives. Hidden goal evaluation
is performed by the evaluator. Writing a desired state into $\Gb$ has no
effect on $\Gw$ or the outcome score.

For a queried area, the visual anchor remains a real source frame, while the
text delta describes local symbolic changes from the episode's initial state.
Repeated queries combine the same recorded visual evidence with the updated
local delta; the delta carries changes caused by counterfactual actions. Belief integration
must preserve this distinction: the visual presence of an object in an old
frame does not override an explicit removal from that area. Likewise,
absence from one observed area does not identify an object's unobserved
destination. Local feedback can correct an assumption without making the
entire kitchen observable.

\subsection{Belief Schema and Update}
\label{app:belief}
The aligned runtime implements $\Gb$ as dictionaries of objects, regions,
containers, and agent position. Table~\ref{tab:belief_schema} describes the
fields used by the evaluated agent.

\begin{table}[htbp]
\centering\small
\caption{Belief-store fields in the aligned runtime.}
\label{tab:belief_schema}
\begin{tabularx}{\linewidth}{lX}
\toprule
Component & Stored information \\
\midrule
Objects & Label-keyed region, confidence, and evidence source; VLM entries can additionally store current state. \\
Regions & Flags recording local or VLM observation. \\
Containers & Known open/closed state and source. \\
Agent & Current region. \\
\bottomrule
\end{tabularx}
\end{table}

Local observations overwrite visible object regions with confidence 1.0.
Successful actions update selected object locations and container states,
after which local observations are integrated again. Explicit removal deltas
delete matching object entries; fused visual observations populate location
and state evidence. The planner view includes up to 30 alphabetically sorted
object entries and the known container states. It does not serialize every
stored state field. No observation-age threshold or confidence-decay policy
is implemented in this aligned baseline. Failure messages support replanning;
they do not guarantee that every erroneous stored fact is removed.

\subsection{Planning and Evaluation Loop}
Algorithm~\ref{alg:interaction} summarizes the protocol rather than prescribing
a particular model's internal reasoning. The controller routes to remembered regions and requests evidence when needed.
The executor may run simulator-generated repairs; the planner can replan
following execution failure. The executor alone applies world mutations.
Persistent and reset conditions use the same world-initialization and reset
protocol; they differ in
the retention of agent-side belief across task groups. Model and execution
budgets are fixed within each controlled comparison.

\newcounter{interactionalgorithm}
\begin{figure}[htbp]
\centering
\begin{minipage}{\linewidth}
\small
\refstepcounter{interactionalgorithm}\label{alg:interaction}
\hrule\vspace{5pt}
\textbf{Algorithm \theinteractionalgorithm. Interaction with a hidden executable world}
\vspace{4pt}\hrule\vspace{5pt}
\textbf{Inputs:} episode world $\Gw_0$; tasks and contracts
$\{(\tau,C_\tau)\}$; rules $\mathcal R$; observation interface $\mathcal O$;
fixed budget; memory condition $m$; task and group reset schedule.
\begin{enumerate}[leftmargin=1.7em,label=\arabic*.,itemsep=2pt,topsep=4pt]
\item Initialize belief $\Gb$ from the initial permitted observation.
\item \textbf{For each} task $\tau$ in episode order:
\begin{enumerate}[leftmargin=1.6em,label=\alph*.,itemsep=2pt,topsep=2pt]
  \item Initialize task record $H_\tau$; retain the world reached by preceding tasks.
  \item Apply $m$ at the configured group boundary; propose a plan from the current group instruction $\tau_g$, $\Gb$, and local context.
  \item \textbf{While} budget remains and the agent has not terminated:
  \begin{enumerate}[leftmargin=1.5em,label=\roman*.,itemsep=2pt,topsep=2pt]
    \item If a new group starts, apply $m$ at that boundary before forming its planning context.
    \item If requested, receive an allowed local observation $o\leftarrow\mathcal O$ and update $\Gb$.
    \item Select the next proposed primitive, using $\Gb$ for any retrieval navigation.
    \item Executor checks primitives on $\Gw$, attempts any supplied repair plan, then attempts the proposed primitive and returns outcomes $f$.
    \item Record execution in $H_\tau$; update $\Gb$ from permitted observations and $f$.
    \item Retain or replan the remaining group actions using updated context, with at most three planner replans per group.
  \end{enumerate}
  \item Evaluator records $C_\tau(\Gw_T,H_\tau)$; carry world state to the next task, resetting only at the next episode.
\end{enumerate}
\end{enumerate}
\vspace{3pt}\hrule
\end{minipage}
\end{figure}

Scoring remains separate from reference-action diagnostics. A terminal
contract reads the final hidden state and accepts only explicitly satisfied
required predicates; missing evidence is conservative failure. An event
contract reads its specified task execution evidence. Equation~\eqref{eq:goal_contract}
keeps these cases distinct, and Appendix~\ref{app:metrics} gives their reported
metrics. A different action order is not itself a reason for terminal failure.
Invalid primitives are rejected locally; terminal success depends on whether
the required outcome holds at the evaluation endpoint.

\section{Evaluation Definitions and Validation}
\label{app:metrics}

The evaluation uses task conditions and execution records supplied by the
benchmark. This appendix defines their scoring semantics, aggregation, and
validation, so that a reported result can be traced to the evidence it tests.

\subsection{Evaluation units and frozen contracts}
\label{app:eval_units}
An action group is a compiled operation, a goal task contains one or more groups associated with an instruction, and an episode orders tasks in a shared world. The aligned comparison fixes 105 task IDs, 18 episode IDs, and 3,441 groups for every planner. The outcome denominator is the task, whereas model calls, action attempts, and VLM queries are process quantities.

Each task has one frozen goal variant with a nonempty \texttt{required} condition list. The schema also supports \texttt{protected} conditions, which, when present, enter the same conjunction and GCR denominator; all 105 enabled variants in the planner comparison have empty protected lists. Terminal conditions test a typed attribute or relation in the final hidden graph. Successful-event conditions specify one or more accepted action types, optional object and target allowlists, and a minimum count. Only executor-accepted task-local events can match them. Missing evidence produces \Unknown{} and fails the binary contract. Reference action sequences and hidden goal predicates are not passed to the planner, and reference event effects are not injected into its trajectory. The agent does receive compiled group instructions, local reachable-action hints, and executor feedback; executor-generated repairs are described in Appendix~\ref{app:feedback_prompt}.

The 105-task set has 34 terminal-state and 71 successful-event contracts. Event-evidence coverage is categorized as direct (56), cross-boundary (7), semantic alias (1), or proxy-limited (7); together with the 34 terminal contracts, 98 tasks form the explicit-evidence sensitivity set. A proxy-limited label records uncovered natural-language components rather than silently treating them as observed. The seven corresponding tasks remain in the fixed primary denominator and are tracked in the explicit-evidence sensitivity analysis.

\subsection{Condition-level attainment}
\label{app:condition_structure}
Equation~\eqref{eq:gcr} averages the truth values of the same frozen conditions used for binary success. It includes both terminal and successful-event tasks and assigns zero to non-completed runs. Because the current schemas enable one variant per task, no cross-variant selection is required. The model-level value is a macro mean over 105 tasks. Since normalization occurs within each task before macro averaging, this construction supports paired cross-task comparison without allowing tasks with longer condition lists to dominate. Terminal and event strata are still reported separately because their conditions encode different kinds of evidence.

\begin{table}[htbp]
\centering\small
\caption{Number of required conditions per task in the frozen 105-task suite. Both types have median two. Counts describe contract structure, not observed success.}
\label{tab:condition_counts}
\begin{tabular}{lrr}
\toprule
Conditions per task & Terminal tasks & Event tasks \\
\midrule
1 & 9 & 11 \\
2 & 13 & 27 \\
3 & 4 & 23 \\
4--5 & 1 & 10 \\
6--10 & 3 & 0 \\
11--29 & 4 & 0 \\
\midrule
Tasks & 34 & 71 \\
Total conditions & 140 & 177 \\
\bottomrule
\end{tabular}
\end{table}

The terminal contracts contain 140 conditions: 61 concern cleanliness, 49 a location target, and 14 contents. The remaining 16 concern mixing, shape or form, power, stirring, wetness, and covering. These are counts of condition definitions, not a breakdown of conditions the planners satisfy. Event contracts contain 177 accepted-event conditions. Table~\ref{tab:condition_counts} shows the number per task: terminal tasks range from one to 29 and event tasks from one to five, with median two in both. GCR normalizes within each task before averaging, so a task with many conditions does not automatically receive more weight. This convention measures relative attainment of the stated requirements; it does not calibrate their difficulty or semantic importance.

\begin{table}[htbp]
\centering\small
\caption{Worked terminal-goal example. The task requires a particular cup to be in the cupboard and the cupboard to be closed.}
\label{tab:goal_example}
\begin{tabularx}{\linewidth}{Xccc}
\toprule
Final execution outcome & Cup in cupboard & Cupboard closed & (Strict-TSR, GCR) \\
\midrule
Cup placed; cupboard left open & True & False & $(0, 0.5)$ \\
Cup placed; cupboard closed & True & True & $(1, 1.0)$ \\
Wrong cup placed; cupboard closed & False & True & $(0, 0.5)$ \\
Cup location unavailable; cupboard closed & Unknown & True & $(0, 0.5)$ \\
\bottomrule
\end{tabularx}
\end{table}

For an event-contract example, a condition can require at least one accepted \texttt{juice}, \texttt{extract}, \texttt{squeeze}, or \texttt{press} event binding an orange half to the juicer cone. Discarding peel or merely moving the cone cannot match this condition. An attempted but rejected squeeze likewise provides no successful-event evidence.

\subsection{Reference coverage and primitive validity}
\label{app:diagnostic_aggregation}
Let $K_\tau=(k_1,\ldots,k_m)$ contain reference occurrences of seven key action types, and let $A_\tau$ be the set of attempted types. Reference coverage is
\begin{equation}
 \RefTCR(\tau)=
 \begin{cases}
  1, & m=0,\\
  \displaystyle\frac{1}{m}\sum_{j=1}^{m}\indicator[k_j\in A_\tau], & m>0.
 \end{cases}
 \label{eq:ref_tcr}
\end{equation}
The seven Ref-TCR categories in Eq.~\eqref{eq:ref_tcr} are \texttt{pick\_up}, \texttt{take\_out}, \texttt{place}, \texttt{put\_in}, \texttt{turn\_on}, \texttt{open}, and \texttt{close}. Five frozen tasks have no reference occurrence in these categories and receive the implemented empty-reference value of one. Across the 105-task manifest there are 1,158 reference key-action occurrences; \texttt{take\_out} has no occurrence. Recorded attempts, including executor repairs, are reduced to attempted-type membership, so repeated submitted attempts cannot increase the score. This metric does not check entity bindings or successful execution.

Task-level action validity divides accepted primitive attempts by $\max(1,N_{\mathrm{attempted}})$. Reported validity is the unweighted mean of the 105 task values; pooled accepted/attempted rates are not used in the paper table. Completed-run coverage is similarly task-macro and equals the fraction of records with terminal status \texttt{completed}. Other terminal statuses, including parse failure, empty plan, and execution block, remain outcome failures.

\begin{table}[htbp]
\centering
\small
\caption{Protocol-aligned evaluation on the same 105 tasks (18 episodes; 3,441 groups per planner). Completed counts runs with terminal status \texttt{completed}. Operational-TSR combines 34 terminal-state and 71 successful-event contracts; Strict-TSR uses the 34 terminal tasks. Other than Completed, values are percentages. Intervals are 95\% scene-stratified episode-cluster bootstrap intervals (10,000 samples).}
\label{tab:planner_results}
\setlength{\tabcolsep}{3.5pt}
\begin{tabular}{lrrrrr}
\toprule
Planner & Completed & Op.-TSR & 95\% CI & Strict-TSR & GCR \\
\midrule
Qwen-Plus & 44/105 & 5.71 & [2.78, 9.20] & 17.65 & 9.29 \\
GPT-5.5 & 59/105 & 5.71 & [2.50, 9.89] & 14.71 & 11.61 \\
Gemini 3.1 Pro & 69/105 & 6.67 & [2.91, 11.76] & 17.65 & 12.31 \\
DeepSeek-V4-Flash & 65/105 & 3.81 & [0.00, 7.29] & 8.82 & 11.48 \\
GLM-5 & 56/105 & 3.81 & [1.02, 7.50] & 11.76 & 7.27 \\
Claude Sonnet 4.6 & 49/105 & 5.71 & [1.74, 11.70] & 14.71 & 9.75 \\
\bottomrule
\end{tabular}
\end{table}

\begin{table}[htbp]
\centering
\small
\caption{Primary task-macro diagnostic means and 95\% scene-stratified episode-cluster bootstrap intervals on the 105-task set (percent). GCR measures graded contract attainment and validity measures accepted-action reliability; both are computed on the same task IDs and provide complementary capability axes. Reference coverage is retained as a process audit in the frozen analysis export.}
\label{tab:diagnostic_intervals}
\setlength{\tabcolsep}{3pt}
\begin{tabular}{lrr}
\toprule
Planner & GCR [95\% CI] & Validity [95\% CI] \\
\midrule
Qwen-Plus & 9.29 [5.90, 13.21] & 57.44 [48.66, 67.86] \\
GPT-5.5 & 11.61 [7.13, 16.22] & 59.34 [55.08, 63.21] \\
Gemini 3.1 Pro & 12.31 [6.30, 20.50] & 54.44 [47.10, 62.40] \\
DeepSeek-V4-Flash & 11.48 [6.90, 16.15] & 52.77 [46.52, 61.82] \\
GLM-5 & 7.27 [3.38, 12.48] & 50.72 [46.90, 55.90] \\
Claude Sonnet 4.6 & 9.75 [4.05, 18.48] & 56.91 [49.29, 66.79] \\
\bottomrule
\end{tabular}
\end{table}

\begin{table}[htbp]
\centering\small
\caption{Outcome profile within each frozen contract type. Values are percentages; brackets give pointwise 95\% episode-cluster intervals. Terminal contracts report Strict-TSR and GCR, while successful-event contracts report their graded GCR signal. The aggregate Operational-TSR remains the fixed-denominator completion baseline over all 105 tasks.}
\label{tab:contract_strata}
\setlength{\tabcolsep}{3pt}
\begin{tabular}{lrrr}
\toprule
& \multicolumn{2}{c}{Terminal state ($n=34$)} & \multicolumn{1}{c}{Successful event ($n=71$)} \\
\cmidrule(lr){2-3}\cmidrule(lr){4-4}
Planner & Strict-TSR & GCR & GCR \\
\midrule
Qwen-Plus & \shortstack{17.65\\{\scriptsize [8.70, 28.57]}} & \shortstack{23.53\\{\scriptsize [15.31, 34.45]}} & \shortstack{2.46\\{\scriptsize [0.00, 5.00]}} \\
GPT-5.5 & \shortstack{14.71\\{\scriptsize [5.88, 25.00]}} & \shortstack{24.25\\{\scriptsize [11.90, 35.80]}} & \shortstack{5.56\\{\scriptsize [2.38, 9.50]}} \\
Gemini 3.1 Pro & \shortstack{17.65\\{\scriptsize [9.09, 27.27]}} & \shortstack{26.40\\{\scriptsize [14.71, 39.39]}} & \shortstack{5.56\\{\scriptsize [1.97, 11.01]}} \\
DeepSeek-V4-Flash & \shortstack{8.82\\{\scriptsize [0.00, 17.86]}} & \shortstack{22.37\\{\scriptsize [11.95, 31.13]}} & \shortstack{6.27\\{\scriptsize [3.13, 9.52]}} \\
GLM-5 & \shortstack{11.76\\{\scriptsize [3.23, 22.73]}} & \shortstack{17.70\\{\scriptsize [6.52, 31.09]}} & \shortstack{2.28\\{\scriptsize [0.86, 4.43]}} \\
Claude Sonnet 4.6 & \shortstack{14.71\\{\scriptsize [4.88, 28.57]}} & \shortstack{21.94\\{\scriptsize [8.45, 40.09]}} & \shortstack{3.92\\{\scriptsize [0.51, 10.09]}} \\
\bottomrule
\end{tabular}
\end{table}

Table~\ref{tab:planner_results} preserves the completion baseline underlying Figure~\ref{fig:planner_overview}, while Table~\ref{tab:diagnostic_intervals} gives uncertainty summaries for the complementary dimensions used to distinguish model behavior. The contract-stratified table reports terminal completion together with graded attainment and uses GCR for the successful-event stratum, where condition-level progress is the informative comparison. Reference coverage remains a process audit rather than a primary ordering dimension because attempted-type membership does not require correct binding, accepted execution, or satisfied task conditions. The endpoint profiles group empty-plan and parse-failure records for display, retain execution blocks separately, and divide completed runs into zero, partial, and full attainment. They partition the same 105 tasks without treating uncompleted runs as evaluated all-false goal states.

\subsection{Same-task reference consistency and evaluator stress tests}
\label{app:evaluator_validation}
\label{app:reference_evidence}
\begin{table}[htbp]
\centering
\small
\caption{Frozen-contract and evaluator audits. The 105-task audit uses exactly the planner-evaluation IDs and no planner outcomes. The 12-task stress test contains 43 predicates and 45 replayed evidence changes.}
\label{tab:contract_audit}
\begin{tabularx}{\linewidth}{Xrr}
\toprule
Check & Scope & Passed \\
\midrule
Frozen reference record satisfies contract & 105 tasks & 105/105 \\
Initial state correctly fails & 105 tasks & 105/105 \\
Predicate-level negative suites pass & 317 suites & 317/317 \\
Reference and action-only replay & 12 tasks & 12/12; 12/12 \\
Partial and wrong-binding negatives & 12 tasks each & 12/12; 12/12 \\
Missing state is unknown and fails & 12 tasks & 12/12 \\
Invalid action leaves state unchanged & 12 tasks & 12/12 \\
Legal independent order swap & 10 eligible & 10/10 \\
\bottomrule
\end{tabularx}
\end{table}

The same-105 audit joins the final task matrix to the 151-task protocol audit frozen before planner evaluation, using exact task IDs. It finds 105 matches and zero unmatched IDs. Frozen reference records pass 105/105 contracts, their recorded task-initial states fail 105/105, and all 317 predicate-level negative suites pass their expected rejection checks. These comprise 140 strict-terminal suites (partial state, wrong binding, and missing state) and 177 operational-event suites (partial events, wrong binding, missing trace, and failed-attempt-only evidence). These are suites, not 317 independent trajectories. The audit includes both contract tiers and uses no planner outcomes.

\paragraph{Where the reference evidence comes from.}
The reference file used by the operational protocol matches the archived continuous-reference input by SHA-256. The reference-generation implementation likewise matches the code hash in its frozen construction protocol. It resets the world once per episode, executes source-derived operations in group order, and retains task start/end graphs, action checks, and repairs. Entity binding uses full world access; preparatory and repair actions are executed through the environment. Thus the reference states originate in continuous execution, rather than separate task resets to desired endpoints.

Subsequent operational-contract preparation selects successful events and, where the reviewed task scope requires it, incorporates evidence across an original task boundary. Eight of the 105 reference records contain such merged action evidence; this record-level count differs from the seven tasks assigned to the cross-boundary coverage category. The contract audit checks the resulting stored graphs and selected events. It is construction evidence under privileged reference access, not an independently scored upper bound under the final planner runner. A same-protocol oracle score would additionally require alignment of runtime dependencies, episode initialization, and the adjusted scoring checkpoints. The relevant starting point is the common episode initial world; recovery from states damaged by an agent's earlier choices is a separate question about failure propagation.

The separate 12-task stress test contains 43 predicates and 45 replayed evidence changes. Reference event replay, action-only transition witnesses, deterministic repeated replay, and invalid-action immutability each pass 12/12. Initial, partial, and wrong-binding cases each fail 12/12; missing state returns \Unknown{} and fails 12/12. Ten tasks permit an independent update swap and both orders succeed in 10/10. These tests demonstrate the stated invariances and conservative failure behavior, not exhaustive semantic equivalence over all objects or action sequences.

\subsection{Bootstrap and paired comparisons}
\label{app:statistics}
For the planner comparison, episodes are grouped by scene. Each bootstrap replicate samples, with replacement, the original number of episodes within every scene and includes all tasks from each sampled episode. Task-macro statistics are then recomputed. We use 10,000 replicates, seed 42, and percentile endpoints. Pairwise TSR replicates reuse the same sampled clusters and subtract the two model rates before taking percentiles.
Table~\ref{tab:pairwise_tsr} reports all paired Operational-TSR differences as the strict completion baseline. The diagnostic comparisons below provide the condition-level and execution dimensions used for model discrimination.

\begin{table}[htbp]
\centering
\small
\caption{Paired Operational-TSR baseline differences in percentage points (row model minus column model). Intervals use the same 10,000 scene-stratified episode-cluster samples; the capability profile in Tables~\ref{tab:gcr_pairwise}--\ref{tab:completion_pairwise} provides the primary model comparison.}
\label{tab:pairwise_tsr}
\begin{tabular}{lrr}
\toprule
Planner pair & Difference (pp) & 95\% CI (pp) \\
\midrule
Qwen-Plus $-$ GPT-5.5 & +0.00 & [-4.12, +3.53] \\
Qwen-Plus $-$ Gemini 3.1 Pro & -0.95 & [-6.06, +2.78] \\
Qwen-Plus $-$ DeepSeek-V4-Flash & +1.90 & [-1.01, +5.43] \\
Qwen-Plus $-$ GLM-5 & +1.90 & [-2.67, +5.75] \\
Qwen-Plus $-$ Claude Sonnet 4.6 & +0.00 & [-6.41, +4.38] \\
GPT-5.5 $-$ Gemini 3.1 Pro & -0.95 & [-5.04, +3.12] \\
GPT-5.5 $-$ DeepSeek-V4-Flash & +1.90 & [-2.36, +7.32] \\
GPT-5.5 $-$ GLM-5 & +1.90 & [+0.00, +4.35] \\
GPT-5.5 $-$ Claude Sonnet 4.6 & +0.00 & [-2.97, +1.98] \\
Gemini 3.1 Pro $-$ DeepSeek-V4-Flash & +2.86 & [-1.12, +8.70] \\
Gemini 3.1 Pro $-$ GLM-5 & +2.86 & [-0.99, +6.49] \\
Gemini 3.1 Pro $-$ Claude Sonnet 4.6 & +0.95 & [-5.06, +5.48] \\
DeepSeek-V4-Flash $-$ GLM-5 & +0.00 & [-5.56, +4.84] \\
DeepSeek-V4-Flash $-$ Claude Sonnet 4.6 & -1.90 & [-9.52, +3.82] \\
GLM-5 $-$ Claude Sonnet 4.6 & -1.90 & [-5.38, +0.00] \\
\bottomrule
\end{tabular}
\end{table}

\paragraph{Paired diagnostic comparisons.}
\label{app:diagnostic_comparisons}
We analyze all 15 model pairs for each of three complementary metrics: GCR, action validity, and completed-run rate. This paired diagnostic analysis retains the 105-task denominator and the same scene-stratified episode draws, including pairing across models and metrics. Tables~\ref{tab:gcr_pairwise}--\ref{tab:completion_pairwise} report all 45 contrasts with pointwise percentile intervals and approximate simultaneous intervals for the joint family.

For contrast $j$, let $\hat\delta_j$ be its observed difference, $\delta_j^{(b)}$ its bootstrap replicate, and $\hat\sigma_j$ the standard deviation across replicates. We take $q$ as the 95th percentile of $\max_j |(\delta_j^{(b)}-\hat\delta_j)/\hat\sigma_j|$ and report $\hat\delta_j\pm q\hat\sigma_j$. Here $q=3.288$. This maximum-deviation construction accounts jointly for the 45 reported contrasts; it uses fixed bootstrap standard errors and remains conditional on the recorded trajectories.

The simultaneous intervals exclude zero for GPT-5.5 versus GLM-5 in validity, and for GPT-5.5, Gemini, and DeepSeek versus Qwen-Plus and Gemini versus Claude in completed-run rate. These metric-specific comparisons characterize the shared planner/controller/repair pipeline, including the actions it supplies. Completed-run rate measures whether the execution loop finishes, including runs with rejected actions or unmet goals; GCR retains graded contract attainment. Together these metrics supply the discriminative profile reported alongside the strict Operational-TSR baseline.

\begin{table}[htbp]
\centering\small
\caption{Goal completion ratio: all 15 paired planner differences, in percentage points (first model minus second). Pointwise intervals use paired episode-cluster percentiles; simultaneous intervals cover the joint family of 45 diagnostic contrasts. Bold differences have simultaneous intervals excluding zero.}
\label{tab:gcr_pairwise}
\setlength{\tabcolsep}{3pt}
\begin{tabular}{lrrr}
\toprule
Planner pair & Difference & Pointwise 95\% CI & Simultaneous 95\% CI \\
\midrule
Qwen-Plus $-$ GPT-5.5 & -2.33 & [-6.12, +1.71] & [-8.86, +4.21] \\
Qwen-Plus $-$ Gemini 3.1 Pro & -3.02 & [-9.89, +1.88] & [-12.85, +6.80] \\
Qwen-Plus $-$ DeepSeek-V4-Flash & -2.20 & [-6.70, +2.44] & [-9.77, +5.37] \\
Qwen-Plus $-$ GLM-5 & +2.01 & [-2.30, +5.63] & [-4.56, +8.59] \\
Qwen-Plus $-$ Claude Sonnet 4.6 & -0.47 & [-7.52, +4.24] & [-10.43, +9.49] \\
GPT-5.5 $-$ Gemini 3.1 Pro & -0.70 & [-7.03, +4.57] & [-10.30, +8.90] \\
GPT-5.5 $-$ DeepSeek-V4-Flash & +0.13 & [-3.11, +3.54] & [-5.38, +5.64] \\
GPT-5.5 $-$ GLM-5 & +4.34 & [-0.04, +8.48] & [-2.84, +11.53] \\
GPT-5.5 $-$ Claude Sonnet 4.6 & +1.86 & [-4.94, +6.93] & [-8.16, +11.88] \\
Gemini 3.1 Pro $-$ DeepSeek-V4-Flash & +0.83 & [-4.78, +8.03] & [-9.86, +11.52] \\
Gemini 3.1 Pro $-$ GLM-5 & +5.04 & [+0.79, +9.72] & [-2.34, +12.42] \\
Gemini 3.1 Pro $-$ Claude Sonnet 4.6 & +2.56 & [-5.09, +8.28] & [-8.63, +13.75] \\
DeepSeek-V4-Flash $-$ GLM-5 & +4.21 & [-1.37, +9.19] & [-4.77, +13.20] \\
DeepSeek-V4-Flash $-$ Claude Sonnet 4.6 & +1.73 & [-6.36, +8.12] & [-10.47, +13.93] \\
GLM-5 $-$ Claude Sonnet 4.6 & -2.48 & [-8.02, +1.57] & [-10.53, +5.56] \\
\bottomrule
\end{tabular}
\end{table}

\begin{table}[htbp]
\centering\small
\caption{Action validity: all 15 paired planner differences, in percentage points (first model minus second). Pointwise intervals use paired episode-cluster percentiles; simultaneous intervals cover the joint family of 45 diagnostic contrasts. Bold differences have simultaneous intervals excluding zero.}
\label{tab:validity_pairwise}
\setlength{\tabcolsep}{3pt}
\begin{tabular}{lrrr}
\toprule
Planner pair & Difference & Pointwise 95\% CI & Simultaneous 95\% CI \\
\midrule
Qwen-Plus $-$ GPT-5.5 & -1.89 & [-8.19, +5.84] & [-13.98, +10.19] \\
Qwen-Plus $-$ Gemini 3.1 Pro & +3.01 & [-1.73, +7.06] & [-4.34, +10.35] \\
Qwen-Plus $-$ DeepSeek-V4-Flash & +4.67 & [-1.32, +10.12] & [-5.06, +14.40] \\
Qwen-Plus $-$ GLM-5 & +6.72 & [-0.14, +13.60] & [-4.98, +18.42] \\
Qwen-Plus $-$ Claude Sonnet 4.6 & +0.54 & [-4.21, +4.49] & [-6.72, +7.79] \\
GPT-5.5 $-$ Gemini 3.1 Pro & +4.90 & [-1.26, +10.22] & [-4.93, +14.73] \\
GPT-5.5 $-$ DeepSeek-V4-Flash & +6.56 & [-1.07, +12.38] & [-4.70, +17.83] \\
GPT-5.5 $-$ GLM-5 & \textbf{+8.61} & [+4.40, +12.14] & [+2.14, +15.09] \\
GPT-5.5 $-$ Claude Sonnet 4.6 & +2.43 & [-4.90, +7.67] & [-8.43, +13.29] \\
Gemini 3.1 Pro $-$ DeepSeek-V4-Flash & +1.66 & [-4.56, +8.18] & [-9.03, +12.35] \\
Gemini 3.1 Pro $-$ GLM-5 & +3.71 & [-1.90, +9.00] & [-5.59, +13.02] \\
Gemini 3.1 Pro $-$ Claude Sonnet 4.6 & -2.47 & [-6.58, +1.51] & [-9.12, +4.18] \\
DeepSeek-V4-Flash $-$ GLM-5 & +2.05 & [-2.15, +6.94] & [-5.55, +9.65] \\
DeepSeek-V4-Flash $-$ Claude Sonnet 4.6 & -4.13 & [-9.45, +0.61] & [-12.55, +4.28] \\
GLM-5 $-$ Claude Sonnet 4.6 & -6.18 & [-11.77, -0.97] & [-15.33, +2.96] \\
\bottomrule
\end{tabular}
\end{table}

\begin{table}[htbp]
\centering\small
\caption{Completed-run rate: all 15 paired planner differences, in percentage points (first model minus second). Pointwise intervals use paired episode-cluster percentiles; simultaneous intervals cover the joint family of 45 diagnostic contrasts. Bold differences have simultaneous intervals excluding zero.}
\label{tab:completion_pairwise}
\setlength{\tabcolsep}{3pt}
\begin{tabular}{lrrr}
\toprule
Planner pair & Difference & Pointwise 95\% CI & Simultaneous 95\% CI \\
\midrule
Qwen-Plus $-$ GPT-5.5 & \textbf{-14.29} & [-20.78, -6.60] & [-26.17, -2.40] \\
Qwen-Plus $-$ Gemini 3.1 Pro & \textbf{-23.81} & [-33.05, -14.78] & [-39.06, -8.55] \\
Qwen-Plus $-$ DeepSeek-V4-Flash & \textbf{-20.00} & [-30.12, -10.08] & [-36.61, -3.39] \\
Qwen-Plus $-$ GLM-5 & -11.43 & [-18.12, -1.30] & [-25.56, +2.71] \\
Qwen-Plus $-$ Claude Sonnet 4.6 & -4.76 & [-12.15, +5.48] & [-19.40, +9.88] \\
GPT-5.5 $-$ Gemini 3.1 Pro & -9.52 & [-17.81, -1.80] & [-22.76, +3.71] \\
GPT-5.5 $-$ DeepSeek-V4-Flash & -5.71 & [-18.48, +5.69] & [-25.65, +14.22] \\
GPT-5.5 $-$ GLM-5 & +2.86 & [-2.50, +10.87] & [-8.27, +13.98] \\
GPT-5.5 $-$ Claude Sonnet 4.6 & +9.52 & [+0.00, +21.21] & [-7.94, +26.98] \\
Gemini 3.1 Pro $-$ DeepSeek-V4-Flash & +3.81 & [-7.06, +13.51] & [-13.67, +21.29] \\
Gemini 3.1 Pro $-$ GLM-5 & +12.38 & [+3.85, +24.14] & [-4.50, +29.26] \\
Gemini 3.1 Pro $-$ Claude Sonnet 4.6 & \textbf{+19.05} & [+11.30, +30.00] & [+3.57, +34.53] \\
DeepSeek-V4-Flash $-$ GLM-5 & +8.57 & [-4.65, +25.49] & [-16.53, +33.67] \\
DeepSeek-V4-Flash $-$ Claude Sonnet 4.6 & +15.24 & [+5.81, +27.27] & [-2.98, +33.46] \\
GLM-5 $-$ Claude Sonnet 4.6 & +6.67 & [-3.33, +16.67] & [-9.74, +23.07] \\
\bottomrule
\end{tabular}
\end{table}

\paragraph{How the strict baseline complements graded attainment.}
Across the 630 model--task records, GCR identifies 78 completed runs with partial attainment that the strict conjunction would collapse into the same binary endpoint as a zero-attainment run. These are repeated evaluations of the same 105 tasks. The strict baseline anchors complete attainment, while GCR preserves graded progress under unchanged contracts. Incomplete runs receive zero by policy, without an observed all-false goal state.

GCR intervals are an offline export from the \texttt{metrics.gcr} field already stored in each immutable task record, using the same resampling code and cohort. Source-record SHA-256 hashes are verified before export. The export computes uncertainty from these frozen scores and terminal statuses; no simulator or model rerun is involved.

The contract-type analysis in Table~\ref{tab:contract_strata} uses the same recorded outcomes. Every replicate first resamples complete episodes within each scene, then computes separate task-macro means for the 34 terminal and 71 event tasks. These subsets occur in 14 and 16 of the 18 episodes, respectively. Samples are shared across models, contract types, and metrics. The plotted intervals are pointwise, not simultaneous bands, and are used to show stratum-specific attainment patterns rather than to define an additional global ranking. More generally, these intervals quantify episode resampling uncertainty for fixed model runs, not variability across new API or model samples. Overall results and previously reported paired intervals remain unchanged.

The Strict-51 memory comparison uses the same episode-cluster procedure and seed, paired by task and episode across conditions. Task metrics are task-weighted; VLM attempts and token use are complete-episode group totals. The 95\% TSR-difference interval crossing zero is evidence of uncertainty, not equivalence or non-inferiority. Provider-attempt reconciliation is kept separate from completed-group token attribution, as detailed in Appendix~\ref{app:runtime_accounting}.

\subsection{Paired memory effects and continuous resource profiles}
\label{app:memory_continuity}
\begin{table}[htbp]
\centering
\small
\caption{Paired Qwen-Plus memory comparison on frozen Strict-51 (51 tasks, 22 episodes, 3,878 groups per condition). Differences are persistent minus per-group reset. Intervals use 10,000 scene-stratified episode-cluster bootstrap samples (seed 42).}
\label{tab:memory_results}
\setlength{\tabcolsep}{3.5pt}
\resizebox{\linewidth}{!}{%
\begin{tabular}{lrrrr}
\toprule
Metric & Persistent & Reset & Difference & 95\% paired CI \\
\midrule
Strict-TSR & 11.76\% & 11.76\% & +0.00 pp & [-6.38, +6.25] pp \\
Action validity & 45.94\% & 41.79\% & +4.15 pp & [+0.11, +7.78] pp \\
VLM query attempts & 95 & 976 & -881 & [-1{,}106{,} -651] \\
Group-accounted tokens & 27.14 M & 23.58 M & +3.56 M & [+1.82, +5.45] M \\
\bottomrule
\end{tabular}}
\end{table}

Table~\ref{tab:memory_results} retains the complete paired effects underlying Figure~\ref{fig:memory_continuity}. The figure adds a descriptive, post-hoc view of existing group records. Within each episode, we sort groups by their source group ID and assign zero-based position $j$ among $n$ groups to quarter $\lfloor4j/n\rfloor$. All 22 episodes have at least four groups and contribute to every quarter. The two conditions have identical group IDs; the four quarters contain 978, 967, 972, and 961 groups per condition. Each resource point divides the total queries or tokens in its quarter by the number of groups, multiplying queries by 100 for readability. Tokens combine planner and VLM group-accounted usage.

For the resource profiles and marginal task-metric intervals, 10,000 bootstrap replicates sample episodes within scenes with replacement (seed 42). Each replicate reuses the same sampled episodes across conditions, quarters, and metrics, retaining all associated records. Resource rates pool groups; validity and Strict-TSR average task values. Shading and whiskers show pointwise percentile intervals, not simultaneous bands. The original paired-difference intervals in Table~\ref{tab:memory_results} are unchanged. Overlap between marginal intervals is not a test of the paired effect.

The public figure input contains minimal projections of 7,756 group records and 102 task records, with each original record's frozen SHA-256. The exporter checks those hashes before retaining group order, resource counts, task metrics, and episode identities. Figure generation reconciles the projections with the published task means and full resource totals. This analysis uses no new model or simulator executions. Relative group progress does not hold task difficulty fixed or manipulate the agent's planning horizon; the first quarter already contains multiple decisions and is not an empty-memory first-action comparison.

\section{Prompts and Agent Implementation Details}
\label{app:implementation}

The planner, executor, belief store, and visual-query module have separate interfaces. The executor reads the complete hidden world; planner-facing messages contain the current group's instruction, allowed action schema, local observation, agent memory, recent feedback, and completed actions. The following configuration and field summaries specify the common interface used in the aligned experiments.

\subsection{Runtime Configuration and Adapter Boundary}
\label{app:runtime_configuration}

\begin{table}[htbp]
\centering
\small
\caption{Planner request identities and optional-reasoning controls. Qwen uses the official DashScope compatible endpoint; the other planners use the same OpenAI-compatible gateway. The gateway resolved the Gemini request to \texttt{gemini-3.1-pro-low}.}
\label{tab:model_settings}
\setlength{\tabcolsep}{4pt}
\begin{tabularx}{\linewidth}{lXX}
\toprule
Paper label & Requested model ID & Request setting \\
\midrule
Qwen-Plus & \texttt{qwen-plus} & \texttt{enable\_thinking=false} \\
GPT-5.5 & \texttt{gpt-5.5} & \texttt{reasoning\_effort=none} \\
Gemini 3.1 Pro & \texttt{gemini-3.1-pro} & \texttt{reasoning\_effort=none} \\
DeepSeek-V4-Flash & \texttt{deepseek-v4-flash} & \texttt{thinking.type=disabled} \\
GLM-5 & \texttt{glm-5} & \texttt{reasoning\_effort=none} \\
Claude Sonnet 4.6 & \texttt{claude-sonnet-4-6} & \texttt{reasoning\_effort=none} \\
\bottomrule
\end{tabularx}
\end{table}

The aligned six-planner run uses temperature 0.1, a maximum of 4,096 output tokens, a 180-second request timeout, and at most four transport attempts (the initial request plus three retries). Each execution group permits at most two visual-model queries, three planner replans in total, and a nominal 80-step loop limit. Qwen-VL-Plus is the visual model for every planner condition. The task order, observation renderer, memory policy, prompt structure, action schema, executor, and evaluator are shared.

Optional reasoning is disabled for every planner through the request control supported by its adapter: \texttt{enable\_thinking=false} for Qwen, \texttt{thinking.type=disabled} for DeepSeek, and \texttt{reasoning\_effort=none} for GPT, Gemini, GLM, and Claude. Qwen uses the official DashScope OpenAI-compatible endpoint; the other five planners use one OpenAI-compatible gateway. These provider-specific request fields and transports implement the shared experiment-level non-thinking policy.

Table~\ref{tab:model_settings} records the requested planner IDs and the gateway-resolved Gemini route. The Gemini response adapter performs only deterministic JSON syntax normalization before the shared schema check: among 18,520 successfully parsed calls, 14,148 are fenced objects with the fence removed, one is a fenced top-level array that is removed and wrapped, 3,144 are native objects, and 1,227 are native top-level arrays wrapped in the expected object field. The adapter does not change action names, arguments, or ordering, and no semantic-repair outcomes are used. Other model adapters likewise map provider response envelopes into the common internal schema. The comparison evaluates deployed API systems under the shared environment-facing protocol, with provider routing and syntax normalization recorded explicitly.

\subsection{Planner Context and Output Schema}
\label{app:prompts}
\label{app:planner_prompt}

The runtime uses \texttt{group\_text} as the goal for each call, not the full
goal-task instruction. Its system message requests strict JSON without
Markdown fences. The user message asks for the shortest remaining plan that
completes the stated operations, discourages repeating completed actions, and
supplies the supported action types and argument conventions.

\begin{table}[htbp]
\centering\small
\caption{Actual field structure assembled by the aligned planner-message function. Local symbolic observations and action hints supplement visual anchors.}
\label{tab:planner_context_fields}
\begin{tabularx}{\linewidth}{lX}
\toprule
Payload component & Contents and truncation \\
\midrule
Local observation & Text, current region, up to 30 visible objects, 20 open storages, 10 active processes, and 50 reachable-action hints. \\
Agent memory & Current region, up to 30 remembered objects (label, region, confidence, source), and known container states. \\
Feedback & Last three execution-failure messages. \\
Completed prefix & Last eight successful actions in this group. \\
Goal and schema & Compiled group instruction, supported action types, and output schema. \\
\bottomrule
\end{tabularx}
\end{table}

The requested output schema contains a \texttt{steps} list:
\begin{verbatim}
{"steps": [
  {"action_type": "<supported action>",
   "object": "<object id, if required>",
   "target": "<destination or region, if required>",
   "steps": 1}
]}
\end{verbatim}
Navigation, opening/closing, and switching use \texttt{target};
\texttt{pick\_up} uses \texttt{object}; retrieval and placement can require
both fields. Parsing and deterministic normalization establish structural
compatibility; the executor separately checks binding and preconditions.
The runtime does not request a separate \texttt{corrected\_action} response
for replanning.

\subsection{State Deltas and Observation Provenance}
\label{app:delta_interface}

A source-video anchor supplies appearance and object evidence for a functional area at its recorded time. The accompanying delta compares the episode's initial symbolic state with its current state in the queried region. These are separate reference points: the image has its own capture time, while the delta's baseline is the initialized world. The belief updater combines the two sources, giving current deltas precedence when they conflict with the fixed image.

The following is a schematic explanation of this interface, not a verbatim
runtime prompt:
\begin{quote}
\small\ttfamily\raggedright
Anchor region: \{queried\_region\}\par
Source frame: \{anchor\_reference\}\par
Local changes since episode initialization:\par
\{local\_state\_deltas\}\par
Describe the objects supported by the image and these changes. Preserve unknown attributes when the evidence does not determine them.
\end{quote}

The delta renderer first selects entities belonging to the queried region, including matching storage and appliance entities. For objects it compares location region, relation, and target; for storage it compares open and locked state; for appliances it compares power and busy state. It reports local additions, removals, and changes to these fields. A removal identifies departure from the queried area without reporting an unobserved destination. This is a region-based symbolic disclosure rule. Planner messages also contain local observations and action hints, while the full-state executor can supply the repairs described below.

\subsection{Repair and Replanning}
\label{app:feedback_prompt}

Before a retrieval action, the controller can insert navigation to an
object's remembered region. It then validates the proposed action. If
validation fails and the simulator returns a \texttt{repair\_plan}, its
primitives are executed through the same environment, followed by another
check and an attempt at the original action. Repair actions are tagged
\texttt{simulator\_feedback\_repair} in the trace. Their accepted and rejected
attempts enter the same task-level action statistics as planner/controller
actions. Validation checks without execution do not enter the attempt count.

After an unsuccessful action, the runtime appends a failure message and may
call the planner with the same group goal and updated context. A nonempty
replacement plan replaces the remaining queue. The counter permits three
such replans per group, despite the configuration key's
\texttt{max\_replan\_per\_action} name. The 80-step bound is checked at the
loop head and during repair; because the original action is subsequently
attempted, it is a nominal loop limit rather than a strict cap after repairs.
An empty queue produces status \texttt{completed}, even if it includes rejected
attempts; a nonempty queue at the limit produces \texttt{execution\_blocked}.
Neither status alone establishes contract satisfaction.

\subsection{Visual Exploration and Belief Updates}
\label{app:vlm_protocol}

Visual queries are triggered for \texttt{pick\_up} or \texttt{take\_out} when
the object is neither locally visible nor located in memory. Location lookup
uses normalized label matching and a minimum confidence of 0.5. Candidate
regions use object--region priors, falling back to the current region and a
fixed kitchen-region list. The image resolver first considers matching region
assets for the episode. For Scene~1 queries only, it also considers the
Scene~1 region-asset directory as a fallback. Catalog assets matching the
requested region have the next priority. If no candidate file is available,
resolution fails. The query record identifies the selected file, its hash,
and its resolution source. Catalog matching is by region; establishing the
image's episode and capture time requires its source metadata. Each group
permits at most two query attempts.

The visual model receives the selected image and queried-region deltas.
Fusion gives those deltas precedence over conflicting anchor evidence.
Memory stores the fused confidence and source, rather than assigning a
universal 0.85 insertion score. Local symbolic observations record visible
object regions with confidence 1.0. Successful retrieval, placement, and
opening/closing update selected object or container entries; local observation
is then integrated again. Consequently, a fine-grained action-derived
location can be overwritten by a region-level observation. These values are
bookkeeping scores, not calibrated probabilities.

The aligned memory implementation has no timestamp-based expiry or age-decay
policy. It removes matching entries for explicit removal deltas and retains
other entries across groups in the persistent condition. Failed-action
messages are available for replanning but do not implement a general
attribute-by-attribute belief correction procedure.

The separate Strict-51 instrumented runtime additionally records field
provenance (source, episode, and group) and diagnostic events. Its
persistent-memory correction counter increments only when an eligible
field changes and the previous write was in a different group of the same
episode. This diagnostic metadata does not implement age-based expiry.

\subsection{Evaluation and Resource Accounting Boundaries}
\label{app:runtime_accounting}

The assigned task set and completed run records are distinct. In the aligned comparison, all 105 assigned tasks remain in the outcome denominator, including parse failures, empty plans, execution blocks, and other non-completed terminal statuses. Primitive validity is determined by executor acceptance, while \RefTCR{} concerns attempts at reference action types. The latter can credit an attempted type whose corresponding primitive fails. Neither diagnostic can be substituted for task-contract satisfaction.

The memory experiment reports VLM query attempts and group-accounted model tokens as separate quantities. Query attempts need not equal successful visual responses. Provider accounting records an additional 3,420 reset-condition tokens from an interrupted and restarted attempt; these are real usage but excluded from completed-group attribution. No monetary-cost or latency estimate is inferred. This distinction matters when memory reduces new observations while increasing the context supplied to model calls.

For reproducible execution, run metadata should bind task identifiers and order, initial-state and contract versions, model endpoints and request settings, prompt and action-registry versions, budgets, retry behavior, and memory configuration. Execution records should retain attempted actions, acceptance results, state transitions, disclosed observations, and model usage fields. These artifacts connect the aggregate outcome table to the exact evaluator and access protocol under which it was produced.

\section{Execution Evidence and Paired Cases}
\label{app:examples}

The dataset's execution interface connects aggregate results to individual
operations. The following analyses use frozen model-run records to examine
action acceptance, retained information, and goal evidence in the same tasks.

\subsection{Step-level execution across planners}
\label{app:trace_analysis}
\begin{table}[htbp]
\centering\small
\caption{Primitive attempts in the frozen six-planner group traces. Repair counts are included in total attempts. These are pooled counts, not task-macro validity percentages. Planner/controller steps include controller-inserted navigation.}
\label{tab:trace_counts}
\setlength{\tabcolsep}{5pt}
\begin{tabular}{lrrrr}
\toprule
Planner & Attempts & Accepted & Rejected & Repair attempts \\
\midrule
Qwen-Plus & 61,302 & 39,292 & 22,010 & 11,763 \\
GPT-5.5 & 43,981 & 28,643 & 15,338 & 7,663 \\
Gemini 3.1 Pro & 46,410 & 28,602 & 17,808 & 8,904 \\
DeepSeek-V4-Flash & 58,739 & 34,075 & 24,664 & 13,421 \\
GLM-5 & 54,948 & 30,536 & 24,412 & 13,874 \\
Claude Sonnet 4.6 & 57,368 & 37,596 & 19,772 & 10,188 \\
\bottomrule
\end{tabular}
\end{table}

We summarize all 3,441 frozen group records for each of the six planners.
Each record's hash is checked against its source index before analysis.
Table~\ref{tab:trace_counts} counts attempts once at the group-trace level,
including simulator-generated repairs. All recorded step statuses in this
cohort are accepted or invalid. The two source tags distinguish
planner/controller execution from executor repairs, not pure model output
from all other actions: the former can include inserted navigation.
These counts describe the amount and composition of execution, whereas the
main validity measure averages per-task fractions. The post-hoc export
retains record identifiers and hashes alongside the counts.

\subsection{Paired task outcomes and their execution evidence}
\label{app:paired_cases}
The 51-task study has 43 failures in both conditions, four successes in both,
two persistent-only successes, and two reset-only successes.
Table~\ref{tab:memory_discordance} includes all four discordant tasks,
covering both directions of the paired difference.
\begin{table}[htbp]
\centering\small
\caption{All four discordant task outcomes in the paired memory study. P: persistent belief; R: per-group reset. Accepted/attempted counts and outcome scores come from task records. An execution block receives zero outcome score even when many attempts were accepted.}
\label{tab:memory_discordance}
\setlength{\tabcolsep}{4pt}
\begin{tabular}{llrrr}
\toprule
Task & Condition & Accepted/attempted & Queries & Success \\
\midrule
Prepare mushrooms & P & 85/122 & 0 & 0 \\
 & R & 26/46 & 4 & 1 \\
Add Squash & P & 16/21 & 0 & 1 \\
 & R & 13/17 & 0 & 0 \\
Wrap tortilla leftovers & P & 23/71 & 0 & 0 \\
 & R & 20/30 & 6 & 1 \\
Drain rice and serve & P & 108/278 & 0 & 1 \\
 & R & 61/212 & 2 & 0 \\
\bottomrule
\end{tabular}
\end{table}

\paragraph{Add Squash.}
In episode \nolinkurl{P02-20240209-231412}, both conditions accept retrieval
of squash in group 2. Their next operations differ. Persistent execution
in group 3 includes an accepted pour of squash into the glass, despite
rejected bottle-opening and closing attempts. At task evaluation, the
glass's recorded contents include squash and the terminal condition is true.
The reset condition instead records three rejected attempts to pick up
squash, followed by an accepted pickup of the glass; no pour is attempted.
Its final evaluation reports the required contents attribute as missing,
so the condition is \Unknown{} and fails. All four groups finish in both
conditions, and their validity fractions are similar (16/21 versus 13/17).
The distinguishing evidence is the executed operation and its goal condition,
not the run-completion flag or a larger validity fraction.

\paragraph{Prepare mushrooms.}
In episode \nolinkurl{P02-20240209-184316}, groups 230--236 contain 85 accepted
attempts out of 122 with persistence, versus 26/46 with reset. However,
persistent group 236 reaches an execution block after 80 attempts, including
74 accepted steps. The task receives zero under the fixed failure policy and
has no goal-evaluation object; it would be incorrect to infer an observed
dirty-mushroom state from that zero. Reset execution completes all seven
groups and its final goal record confirms the mushroom's clean state.
This case separates frequent local acceptance from finishing an operation
and obtaining an evaluated outcome.

The paired records localize behavioral divergence to concrete operations
and outcome checks. The local export includes selected group traces,
terminal condition evaluations, task metrics, and source hashes. These
observations characterize the two retention conditions; attributing an
outcome to a particular belief entry would require an intervention on that entry.

\subsection{A traceable execution excerpt}
Table~\ref{tab:logged_sequence} summarizes Qwen-Plus with persistent memory
in Scene~1, groups 1--7 of episode \nolinkurl{P01-20240202-110250}.
The action and memory excerpts come from frozen group records, with their
source hashes retained in \texttt{data/content\_audit.json}. Rows summarize
recorded operations; they do not reconstruct unrecorded hidden-state attributes.

\begin{table}[htbp]
\centering\small
\caption{Logged coffee-preparation excerpt. All seven group records have
runtime status \texttt{completed}; this status is not a claim of successful
goal attainment. Accepted actions and stored beliefs are distinct evidence.}
\label{tab:logged_sequence}
\begin{tabularx}{\linewidth}{rXX}
\toprule
Group & Instruction and execution excerpt & Recorded belief evidence \\
\midrule
1 & Retrieve a mug: \texttt{open(cupboard)} accepted. & Mug region recorded as cupboard from local observation; no retrieval action in this group. \\
2 & Place mug under nozzle: pickup, navigation, and placement accepted. & Mug region subsequently recorded as countertop, not a fine-grained nozzle relation. \\
3 & Switch on machine and insert capsule: switch, pickup, and placement accepted. & Capsule region recorded as countertop after execution. \\
4 & Retrieve milk from fridge: navigation accepted. & Only navigation is recorded; milk remains a retained countertop belief. \\
5 & Move frother: placement rejected, subsequent move accepted. & Frother region recorded as countertop. \\
6 & Open, pour, close milk: pickup and pouring accepted; four opening attempts and one closing attempt rejected. & Mixed acceptance despite completed runtime status. \\
7 & Turn on frother: four attempts rejected. & Queue exhaustion still yields completed status. \\
\bottomrule
\end{tabularx}
\end{table}

These records illustrate why neither a completed group nor a reference action
type establishes task success. They also expose the granularity of the
baseline memory: local observation can overwrite a placement-derived target
with a coarser region. Task contracts are evaluated separately on the
appropriate hidden state or accepted-event record.

\subsection{Executor repair versus planner actions}
In group 11 of the same episode and condition, the trace records
\texttt{go\_to(drawer)} as an accepted
\texttt{simulator\_feedback\_repair}. Four subsequent
\texttt{pick\_up(scissor)} attempts are rejected. Later planner-tagged steps
navigate to the countertop, pick up \texttt{mesh}, navigate again, and cut
\texttt{mesh}, all with accepted status. The source tags make the executor's contribution to recovery explicit,
while subsequent steps show how execution continues. Goal evaluation remains
separate from the acceptance of these individual operations.

\subsection{Re-observing a static anchor after a state change}
\label{app:anchor_example}
Repeated visual queries preserve the source frame and update the disclosed
local changes. An explicit removal delta takes precedence over the object's
presence in that frame, while unobserved destinations remain unknown.
This rule gives observation evidence a consistent temporal interpretation
across interactions (Eq.~\eqref{eq:delta_observation}).

\subsection{Memory persistence and stale facts}
The recorded coffee excerpt shows how local observations can replace a
fine-grained placement target with a coarser region. Such updates make belief
granularity relevant to later execution. The paired experiment measures the
net effects of persistence on the frozen Strict-51 cohort. The logger records
351,114 cross-group memory hits with persistence versus zero with per-group reset;
repeated region queries fall from 370 to two. These counters accompany the
continuous resource profiles in Figure~\ref{fig:memory_continuity}. The 632 persistent-memory
field revisions count eligible changes whose previous provenance belongs
to an earlier group of the same episode. Such changes do not by themselves
prove that the old belief was erroneous when stored. The separate anchor--delta
conflict counts are three with persistence and 218 with reset; neither counter
measures every stale belief or every failed action.

\section{Artifact Scope and Responsible Use}
\label{app:artifacts}

\paragraph{Resource and release scope.}
\kb{} comprises source-linked operations, episode worlds, transition records, and task conditions. We will release the compiler, action rules, executor, observation interface, initial worlds, source mappings, and frozen contracts needed to run new agents on the two reported evaluation suites. Their scope is given in Table~\ref{tab:construction_scope}; the source HD-EPIC scale is not a claim of full-corpus benchmark coverage. The release will distinguish precompiled task use from rebuilding assets with authorized source-data access. Raw videos remain governed by HD-EPIC terms.

\begin{table}[htbp]
\centering\small
\caption{Benchmark components and the scope of the accompanying manuscript package. Interactive components belong to the separate benchmark release; the manuscript package reproduces analyses from recorded evidence.}
\label{tab:artifact_scope}
\begin{tabularx}{\linewidth}{@{}lXX@{}}
\toprule
Component & Research use & Manuscript package \\
\midrule
Worlds and action rules & Execute new proposals and inspect state changes & Method and interface specifications \\
Source and task mappings & Trace task objects and operations to activity evidence & Cohort identities and the Add Squash example \\
Contracts and scorer & Evaluate states and accepted events & Definitions, audits, and condition inventory \\
Agent and execution records & Examine decisions, observations, and results & Frozen metric exports and selected/minimal trace projections \\
Analysis scripts & Reproduce tables, figures, and uncertainty & Included; no provider access required \\
Video assets & Supply source appearance and context & Access through the source dataset \\
\bottomrule
\end{tabularx}
\end{table}

\paragraph{Using the benchmark data.}
Episode worlds and ordered tasks provide activity contexts for running an
agent. Source-step links let a researcher inspect how an object or operation
entered the dataset. Transition records expose the bindings, preconditions,
and effects of executable operations. During evaluation, action traces and
frozen task conditions show how a proposed solution unfolds and whether it
achieves the required result. The same world rules can be used to compare
planning or information-retention policies, as in the paired memory study.

\paragraph{Manuscript reproduction.}
The manuscript package contains anonymized frozen exports, task-level metrics, configurations, audit summaries, SHA-256 manifests, regeneration scripts, figures, and \LaTeX{} source. It is a complete input for reproducing the reported analyses, with the narrower scope in Table~\ref{tab:artifact_scope}. The importer verifies all 630 planner task-record hashes before exporting stored GCR values; it neither calls a model nor reruns the simulator. Re-execution still requires benchmark code, initial worlds, raw action records, and provider access.

Compiler records include pre- and post-operation graphs. Model-run traces
retain attempts, source tags, execution statuses, group-end belief, and task
evaluations, but not every intermediate world snapshot. The execution
analysis verifies 20,646 group-record hashes; paired examples are checked
against the original export manifest. Portable projections of all 7,756
paired group records and 102 task records reproduce the continuous resource
profiles and task-metric intervals. The condition inventory retains counts,
families, and a source-file hash for all 105 frozen schemas; together with
630 task-level outcomes, it reproduces the contract-type results and
condition distribution. The reference-provenance summary links reference
generation, its frozen protocol, and the input used for goal preparation
through matching hashes. The diagnostic analysis additionally reproduces all 45 paired contrasts across GCR, action validity, and completed-run rate with joint simultaneous intervals; these are the primary comparison dimensions alongside the strict Operational-TSR baseline. These analyses use existing records without model
or simulator calls.

\paragraph{Versions and evaluation identity.}
A run is identified by its tasks, initial worlds, contract and simulator versions, model configuration, and observation/memory settings. The 105-task comparison, paired memory study, and auxiliary diagnostics are separate settings whose denominators must remain attached. Manuscript copies remove machine-local path prefixes while retaining source-relative identifiers and hashes.

\paragraph{Computation and model usage.}
Compilation checks and symbolic replay are CPU-based; planner and visual components use external APIs. We report model identities and measured resources without inferring provider hardware or monetary cost. LLM-assisted grouping and semantic review remain constrained by source checks, schemas, and executable replay.

\paragraph{Licensing.}
Access to and redistribution of source data remain governed by HD-EPIC terms.
The manuscript package grants no right to redistribute third-party videos
or assets. Reuse must respect the applicable licenses; primary datasets and
software dependencies are cited.

\FloatBarrier
\end{document}